\documentclass[conference]{IEEEtran}
\IEEEoverridecommandlockouts
\usepackage{times}

\usepackage[numbers]{natbib}
\usepackage{multicol}
\usepackage[bookmarks=true]{hyperref}
\usepackage[utf8]{inputenc}
\usepackage[table]{xcolor}
\usepackage{graphicx} 
\usepackage{caption}
\usepackage{textgreek}
\usepackage{amsfonts}
\usepackage{amsmath}
\usepackage{amssymb}
\usepackage{subfig}
\usepackage{bbm}
\usepackage{tikz}
\usetikzlibrary{shapes.geometric, arrows.meta, positioning, fit, backgrounds, calc}
\usepackage{booktabs}
\usepackage{multirow}
\usepackage{algorithm}
\usepackage{algpseudocode}

\usepackage{xcolor}

\definecolor{my_pink}{HTML}{be0027}

\begin{document}
\title{Learning while Deploying: Fleet-Scale Reinforcement Learning for Generalist Robot Policies}
\author{
Yi Wang$^{1,2}$ \quad
Xinchen Li$^{2}$ \quad
Pengwei Xie$^{2}$ \quad
Pu Yang$^{2}$ \quad
Buqing Nie$^{2}$ \\
Yunuo Cai$^{1,2}$ \quad
Qinglin Zhang$^{2}$ \quad
Chendi Qu$^{2}$ \quad
Jeffrey Wu$^{3}$ \quad
Jianheng Song$^{2}$ \\
Xinlin Ren$^{2}$ \quad
Jingshun Huang$^{1,2}$ \quad
Mingjie Pan$^{1,2}$ \quad
Siyuan Feng$^{2}$ \quad
Zhi Chen$^{2}$ \quad
Jianlan Luo$^{1,2\dagger}$
\thanks{$\dagger$ Corresponding author.} \\
$^1$Shanghai Innovation Institute. $^2$AGIBOT Finch.
$^3$Columbia University.\\

    \\
\href{https://learning-while-deploying.github.io/}{\textcolor{my_pink}{https://learning-while-deploying.github.io/}
}
}




%

\maketitle

\begin{abstract}

Generalist robot policies increasingly benefit from large-scale pretraining, but offline data alone is insufficient for robust real-world deployment. Deployed robots encounter distribution shifts, long-tail failures, task variations, and human correction opportunities that fixed demonstration datasets cannot fully capture. We present \emph{Learning While Deploying} (LWD), a fleet-scale offline-to-online reinforcement learning framework for continual post-training of generalist Vision-Language-Action (VLA) policies. Starting from a pretrained VLA policy, LWD closes the loop between deployment, shared physical experience, policy improvement, and redeployment by using autonomous rollouts and human interventions collected across a robot fleet. To stabilize learning from heterogeneous, sparse-reward fleet data, LWD combines \emph{Distributional Implicit Value Learning} (DIVL) for robust value estimation with Q-learning via \emph{Adjoint Matching} (QAM) for policy extraction in flow-based VLA action generators. We validate LWD on a fleet of 16 dual-arm robots across eight real-world manipulation tasks, including semantic grocery restocking and 3--5 minute long-horizon tasks. A single generalist policy improves as fleet experience accumulates, reaching an average success rate of 95\%, with the largest gains on long-horizon tasks.

\end{abstract}

\IEEEpeerreviewmaketitle

\begin{figure}
    \includegraphics[width=\linewidth]{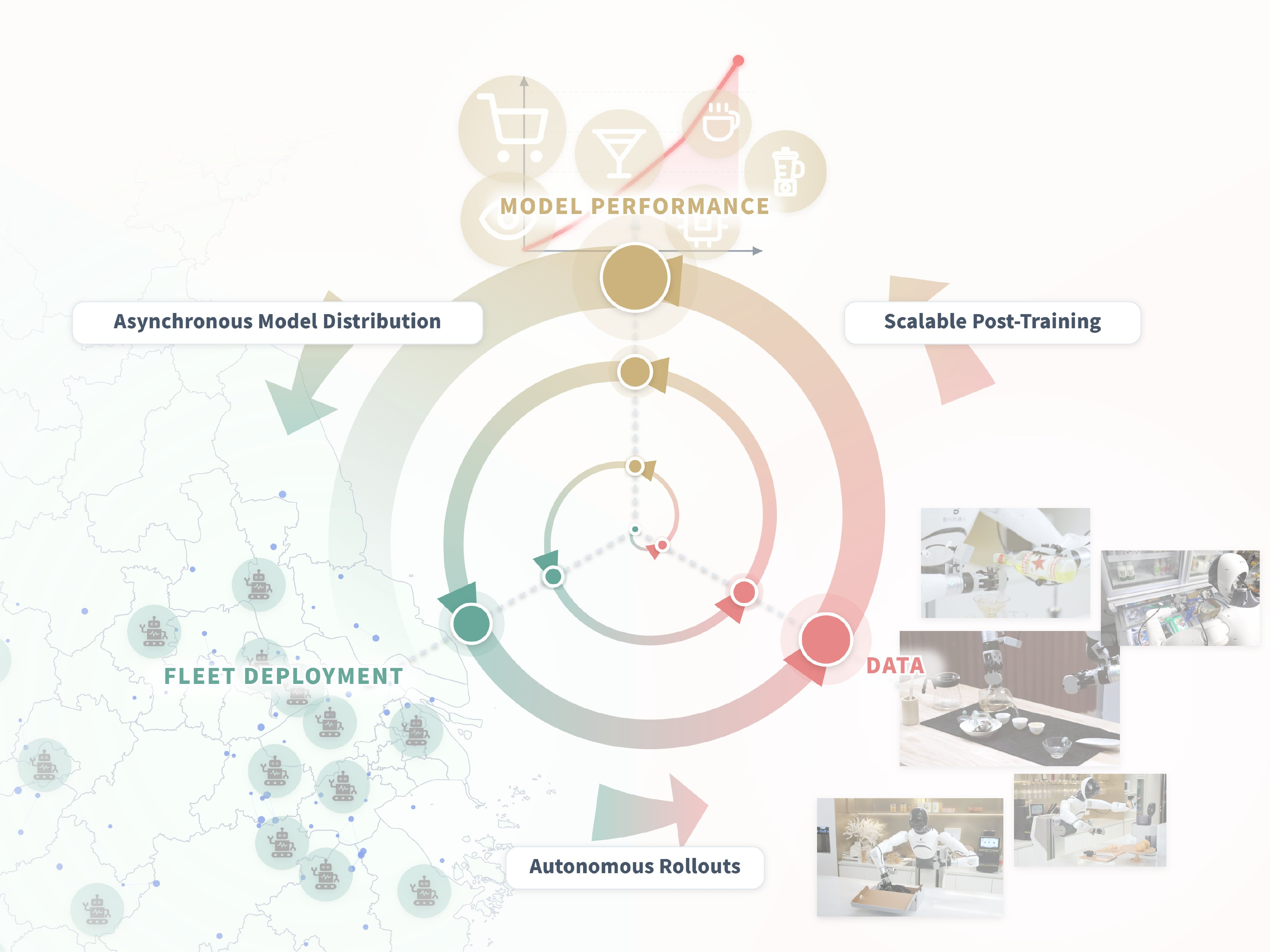}
    \captionof{figure}{\textbf{Learning While Deploying (LWD): Fleet-scale Reinforcement Learning for Generalist Robot Policies.} A pretrained Vision-Language-Action (VLA) model is first initialized with \textit{human-collected} offline data. The data flywheel then spins up. The model is deployed across diverse real-world robot tasks and \textit{autonomously} collects online interaction data. This online data is mixed with the offline replay buffer to update the model, which is then re-deployed for further data collection. }
    \label{fig:teaser}
\end{figure}
\bstctlcite{setting}
\section{Introduction}
Deploying general-purpose robots in the real world requires \emph{high-performance generalist} policies: policies that can reliably complete a broad range of tasks across diverse objects, environments, user instructions, and operating conditions. Recent Vision-Language-Action (VLA) policies~\citep{rt1,rt2,octo,openvla,pi0,pi05} provide a strong foundation by acquiring broad competence from large offline robot datasets. However, offline pretraining alone does not make a policy deployment-ready. Real-world deployment is not a fixed test distribution: as robots are used across more homes, stores, workspaces, and users, they encounter new tasks, object instances, configurations, preferences, and rare failure modes beyond the coverage of pretraining data. Obtaining high performance therefore requires policies that continue to improve from deployment experience, so that adaptation scales with the data generated by use.

This perspective recasts deployment from an endpoint of training into a source of continual policy improvement. 
Realizing this form of continual improvement requires deployment experience that is both broad and continuously updated.
For a generalist robot policy, the most valuable deployment experience is naturally collected at fleet scale. Any individual robot samples only a small portion of the deployed distribution, whereas a fleet spans diverse tasks, environments, objects, and user instructions, producing heterogeneous experience that includes successes, failures, recoveries, partial progress, rare edge cases, and occasional human interventions. Aggregating this physical experience through a shared policy creates a closed-loop data flywheel: deployed robots generate experience on the target deployment distribution, the shared policy improves from the aggregated data, and the improved policy is redeployed to collect broader and more informative experience.

We refer to this setting as \emph{Learning While Deploying} (LWD): continual policy improvement driven by the accumulated real-world autonomous experience of a deployed robot fleet.
Turning this data flywheel into a learning algorithm, however, requires a training objective that can improve from the outcomes of autonomous interaction, rather than treating deployment data for pure imitation signal. Interactive imitation-learning methods~\citep{kelly2019hg} can incorporate expert demonstrations, corrections, and interventions during deployment, but they treat deployment primarily as a source of action labels for supervised learning. As a result, they use only part of the available experience and lack a principled mechanism for leveraging autonomous trials that contain successes, failures, recoveries, partial progress, and task rewards. 
Reinforcement Learning (RL) in principle provides such a mechanism by optimizing policy behavior from task outcomes and policy experience~\citep{Qlearning,td3,ddpg,sac}.
Yet existing RL approaches for robotics are often limited to small-scale, short-horizon, or task-specific settings, and frequently specialize a pretrained generalist policy to a narrow task~\citep{rl100,grrl,conrft}. 
A scalable method for post-training end-to-end VLA policies from fleet deployment experience while preserving their generality remains an open problem.

Addressing this gap requires an RL algorithm for LWD that is compatible with pretrained VLA policies, can learn from large offline and off-policy datasets, and can adapt rapidly as new deployment data streams in. These requirements stress both components of an RL method. 
Value learning must produce reliable estimates from heterogeneous off-policy data with sparse rewards and rare high-return trajectories. Policy extraction must turn the learned values into better actions from a large generative VLA policy without destabilizing the model.

Prior work addresses these requirements only in part. \citet{pi06} combines offline value learning with iterative offline RL, but the procedure is slow and does not directly use action gradients from the learned value function.~\citet{serl,hilserl} show that online RL can learn challenging robotic manipulation tasks within a short period through real-world interaction, but train task-specific policies from scratch rather than improve a pretrained generalist policy. On-policy VLA finetuning methods~\citep{lu2025vla,tan2025interactive,chen2025pirl,flowgrpo,reinflow} update pretrained policies directly from online rollouts, but are not designed to efficiently reuse large offline or off-policy deployment buffers. They also do not learn an explicit action-value critic, and therefore cannot use action-space critic gradients to guide policy improvement. Together, these limitations motivate an offline-to-online RL approach that can reuse heterogeneous deployment data while stably improving a pretrained generative VLA policy.

We present Fleet-Scale Offline-to-Online RL, an offline-to-online framework for post-training end-to-end VLA policies in a large-scale real-world deployment system. The framework couples two pieces: distributional value learning from offline and autonomous deployment experience, and stable policy extraction that transfers value improvement into a flow-based VLA policy.

For value learning, we introduce Distributional Implicit Value Learning (DIVL). DIVL builds on the value-learning component of Implicit Q-Learning~\citep{kostrikov2021offline}, but replaces scalar expectile value regression with a distributional value model. 
This choice is important in the setting of fleet deployment since robots collect diverse data asynchronously under a variety of conditions.
As a result, the return associated with the same state-action pair can be multi-modal and heavy-tailed. 
A scalar critic may collapse these outcomes into an average value and obscure rare but reproducible successes, whereas a distributional critic can preserve these high-return modes.
DIVL therefore learns multi-step return distributions while retaining the in-support policy improvement property of implicit value learning. This yields a stable learning signal from large off-policy deployment buffers without requiring the policy to query out-of-distribution actions.

For policy extraction, we adopt Q-learning with Adjoint Matching (QAM)~\citep{domingo2024adjoint,li2026q}. The critic provides useful action gradients, but backpropagating them through the full multi-step denoising process of a flow policy is unstable and expensive. QAM converts the critic gradient at the denoised action into step-wise supervision for the flow model. This gives a stable way to update the VLA policy from the learned value function while preserving the expressivity of generative action modeling.

The full system has two stages: offline pretraining on a mixture of data from diverse sources, followed by rapid online finetuning with deployment data. 
Both stages optimize the same RL objective, which mitigates a common offline-to-online mismatch: offline critics can become overly conservative and poorly calibrated for subsequent online finetuning, while online improvement depends on extrapolating values to newly visited actions~\citep{nakamoto2023cal}.
We instantiate it on a fleet of 16 dual-arm robots across eight manipulation tasks. These include long-horizon precision tasks, such as brewing Gongfu tea, making cocktails, and making fruit juice, which typically require 3--5 minute executions, as well as shorter-horizon tasks that require semantic generalization, such as restocking diverse items in grocery stores.
A single generalist policy trained with LWD improves as online fleet experience accumulates. It substantially improves over the pretrained model, reaches an average success rate of $0.95$ across all tasks, and outperforms relevant baselines by large margins. 
The performance gap is especially pronounced on long-horizon tasks, where RL can propagate rewards through multi-step dynamic programming and stitch together value estimates across partial progress, while imitation-learning methods suffer more severely from compounding errors.
This LWD procedure typically requires only a few hours of real-world interaction.

Our main contribution is a fleet-scale offline-to-online RL system for post-training generalist robot policies in real-world deployment. Algorithmically, LWD combines distributional implicit value learning with QAM-based policy extraction and uses the same RL objective across offline pretraining and online finetuning. Systemically, it enables a distributed robot fleet to aggregate physical interaction experience and autonomously improve a shared VLA policy.
To the best of our knowledge, LWD is among the first real-world RL systems to close this offline-to-online improvement loop for generalist robot policies. More broadly, it provides a concrete step toward deploying general-purpose robots at scale: fleet-scale deployment can itself become a source of training data, creating a data flywheel in which deploying more robots improves the shared policy and, in turn, future deployment.

\section{Related Work}

LWD is a post-training framework for generalist robot policies, instantiated as a distributed large-scale RL system deployed in real-world settings. 
Accordingly, we survey prior work in the following areas.

\subsection{Post-Training of Robot Generalist Policies}
Robot generalist policies, including VLA models, acquire broad capabilities through large-scale pre-training on diverse multi-modal data~\citep{rt2, octo, openvla, pi05}.
To adapt these policies to downstream deployments, recent work has explored several post-training strategies~\citep{zhang2024grape,chen2025pirl,rl100,pi06,zang2025rlinf,liu2025can}.
One direction studies offline post-training, where policies are improved using previously collected rollouts~\citep{zhang2024grape,pi06,xu2024rldg}.
$\pi^*_{0.6}$ combines offline value learning with iterative offline RL, achieving substantial gains on individual real-world tasks~\citep{pi06}.
RLDG uses specialist RL to generate data for policy distillation, providing another way to incorporate RL supervision~\citep{xu2024rldg}.
However, offline-only post-training follows a collect-train-deploy cycle and cannot immediately incorporate experience gathered during deployment, making adaptation to distribution shifts slow~\citep{pi06,xu2024rldg}.
LWD instead updates the policy during deployment, allowing newly collected experience to correct such shifts quickly.

Another line of work performs post-training with online RL, including VLA-RL~\citep{lu2025vla} and RIPT~\citep{tan2025interactive}, achieving strong improvements for specialist policies in simulated tasks~\citep{behavior1k,maniskill,libero,robotwin,chen2025pirl,zang2026rlinf,jiang2026wovr}.
However, these methods typically rely on on-policy data collection, which can be sample-inefficient and costly for real-world robots~\citep{chen2025pirl,li2025simplevla}.
In contrast, LWD learns from large offline datasets together with off-policy online replay, improving the practicality of real-world post-training.

Recent methods also combine offline and online phases: offline pretraining on rollout datasets followed by online refinement through real-time interactions~\citep{grrl,conrft,rl100}.
However, prior methods typically learn specialist policies tailored to individual tasks, limiting generalization across diverse deployments~\citep{grrl,rl100}.

LWD is fundamentally different from these works: it performs offline-to-online post-training for a generalist robot policy rather than learning task-specific specialists.
This enables scalable post-training of a single policy across multiple real-world tasks, including long-horizon tasks with sparse rewards.

\subsection{Offline-to-Online Reinforcement Learning}
Offline-to-online RL pretrains on diverse offline data and refines continuously through online interactions~\citep{park2025flow,li2026q,kununi,lee2022offline,rl100,conrft,xu2024rldg,agarwal2022reincarnating,ball2023efficient}.  
\citet{serl,hilserl} utilize a small number of human demonstrations to seed policy learning and then specialized a single robotic skill through real-world interaction. 
However, LWD differs from this method in that it post-trains a shared generalist VLA policy across multiple tasks, combines offline and online replay in one learning loop, and operates under distributed fleet-scale deployment.
Recent studies use different policy-extraction mechanisms to reuse offline data during online improvement~\citep{li2026q,nair2020awac,song2022hybrid,wagenmaker2025steering}. 
\citet{wagenmaker2025steering} present DSRL, which adapts pretrained diffusion policies via RL on latent-noise space for sample-efficient online-to-offline improvement.
\citet{li2026q} introduce QAM, using critic gradients to improve flow-based policies through adjoint matching, achieving stable training from scratch in simulation. 
However, prior approaches have not been validated for stable, fleet-scale post-training of generalist VLA policies. LWD addresses this and adopts QAM to enable offline-to-online RL through large-scale real-world deployments.

Recent robotic post-training methods incorporate offline-to-online RL to improve policies~\citep{grrl,rl100,conrft,xu2024rldg}. 
However, they typically focus on task-specific policies with inconsistent training objectives across offline-to-online stages, and operate at limited deployment scale.
In contrast, LWD trains a generalist policy across diverse tasks through fleet-scale offline-to-online RL.
It adopts a unified training method in offline and online stages, enhancing training stability and scalability.

\subsection{Large-Scale Robotic RL Systems}
Large-scale robotic RL systems improve robot policies by aggregating experience from distributed actors and training centralized learners, enabling policy improvement beyond what can be achieved with isolated task-level data collection~\citep{kalashnikov2018qtopt,kalashnikov2021mtopt,lee2023pi,pan2026sop,bousmalis2023robocat,espeholt2018impala,herzog2023deep}.~\citet{kalashnikov2018qtopt,kalashnikov2021mtopt} demonstrate that off-policy RL can be scaled from vision-based grasping to multi-task manipulation through asynchronous robot data collection and centralized Q-function optimization.
While these systems focus primarily on short-horizon manipulation and learn policies largely from scratch, LWD post-trains a pretrained generalist VLA policy across diverse real-world tasks, including long-horizon manipulation.~\citet{bousmalis2023robocat} and~\citet{herzog2023deep} further study learning from large-scale robot experience, but the former relies on behavior cloning while the latter targets task-specific RL for waste sorting.
Most recently, SOP~\citet{pan2026sop} formalizes the system substrate for scalable online post-training of VLA policies, coupling a distributed robot fleet with a centralized cloud learner and asynchronous policy synchronization.
Building on this deployment substrate, LWD instantiates the learning algorithm with offline-to-online RL: it jointly leverages prior offline data and newly collected fleet experience to improve a single generalist policy across long-horizon real-world tasks.

This distinction shifts the contribution from distributed execution alone to an RL-driven data flywheel, where large-scale deployment continually supplies experience for policy improvement.

\section{Preliminaries}
\label{sec:prelim}

\subsection{Problem Setting and Notation}

We formulate robot control as a Markov decision process
$\mathcal{M}=(\mathcal{S},\mathcal{A},\mathcal{T},r,\gamma)$, where $\gamma\in(0,1]$ is the discount factor. We consider a set of tasks indexed by $k\in\mathcal{K}$. Each state $s = (o,\ell_k) \in \mathcal{S}$ consists of a robot observation $o$ and a language instruction $\ell_k$ specifying task $k$. For long-horizon tasks, $\ell_k$ is a high-level command, such as `Make Tea' rather than a sequence of low-level subtask instructions.
In our setting, we use sparse binary rewards, with $r=1$ only when an episode terminates successfully and $r=0$ otherwise.

LWD trains one shared generalist VLA policy across all tasks. At time $t$, the policy takes the state $s_t$ as input and outputs an action chunk
\begin{equation}
    \mathbf{a}_t \equiv \mathbf{a}_{t:t+H}
    =
    [a_t,a_{t+1},\ldots,a_{t+H-1}]
    \sim \pi_\theta(\cdot\mid s_t),
\end{equation}
which is executed before replanning. The corresponding chunk reward is
\begin{equation}
    \mathbf{r}_t \equiv \mathbf{r}_{t:t+H}
    =
    \sum_{i=0}^{H-1}\gamma^i r_{t+i}.
\end{equation}
Thus, mixed-task replay samples are written abstractly as
$(s_t,\mathbf{a}_t,\mathbf{r}_t,s_{t+H})\sim\mathcal{D}$, where $\mathcal{D}$ denotes the replay distribution. In the offline stage, $\mathcal{D}$ is induced by samples from $\mathcal{B}_{\mathrm{off}}$; in the online stage, it is induced by mixed replays from $\mathcal{B}_{\mathrm{off}}\cup\mathcal{B}_{\mathrm{on}}$.
Throughout the method, the generalist policy and critic operate on action chunks.

\subsection{Implicit Q-Learning}
Implicit Q-Learning (IQL)~\citep{kostrikov2021offline} avoids explicit action maximization by fitting a \textit{scalar} state-value function to a high \textit{expectile} of dataset action-values. Using the chunk notation from above, with $\mathbf{a}_t$ and $\mathbf{r}_t$, IQL fits
\begin{equation} \label{eq:loss_iql}
    \mathcal{L}_V^{\mathrm{IQL}}(\psi) = \mathbb{E}_{\mathcal{D}} \left[ \rho_{\tau,2}(Q_{\bar{\phi}}(s_t,\mathbf{a}_t)-V_\psi^{\mathrm{IQL}}(s_t)) \right], 
\end{equation}
where
\begin{equation}
    \rho_{\tau,2}(u) = \left|\tau-\mathbb{I}(u<0)\right|u^2 ,
\end{equation}
and $Q_{\bar{\phi}}$ denotes the target network, whose parameters are updated by exponential moving average. The critic $Q_{\phi}$ is trained with the value-based TD target
\begin{equation}
y_t^{\mathrm{IQL}} = \mathbf{r}_t + \gamma^HV_\psi^{\mathrm{IQL}}(s_{t+H}),
\end{equation}
using
\begin{equation}
\mathcal{L}_Q^{\mathrm{IQL}}(\phi)
=
\mathbb{E}_{\mathcal{D}}
\left[
\left(
Q_\phi(s_t,\mathbf{a}_t)
-
y_t^{\mathrm{IQL}}
\right)^2
\right].
\end{equation}
For $\tau>1/2$, this value estimate is biased toward higher-valued dataset actions, giving an implicit improvement target without a $\max_{\mathbf{a}}Q(s,\mathbf{a})$ backup. In LWD, we retain this \textit{asymmetric bootstrap} principle, but replace scalar expectile value regression with a \textit{distributional} value model and \textit{quantile}-based value extraction.
\subsection{Flow Matching and Q-learning with Adjoint Matching}
\label{subsec:qam}

Flow Matching (FM)~\citep{lipman2022flow} represents a generative policy as a time-dependent vector field. Given a data action chunk $\mathbf{a}^1 = \mathbf{a}$ and Gaussian noise $\mathbf{a}^0\sim\mathcal{N}(0,I)$, FM defines the interpolation
\begin{equation}
\label{eq:fm-interpolation}
    \mathbf{a}^w = (1-w)\mathbf{a}^0 + w\mathbf{a}^1,
    \qquad w\in[0,1],
\end{equation}
and trains a conditional vector field $f_\theta(s,\mathbf{a}^w,w)$ to match the velocity $\mathbf{a}^1-\mathbf{a}^0$. Flow-based VLA policies use this construction as an action-generation head~\citep{pi0,pi05}.

For policy extraction, a flow policy must be optimized through a multi-step generation process, making direct critic backpropagation costly and potentially unstable. Q-learning with Adjoint Matching (QAM)~\citep{li2026q} addresses this problem by combining TD critic learning with an adjoint-matching policy update. Given a pretrained reference flow $f_\beta$ and a critic $Q_\phi$, QAM defines the KL-regularized improvement target
\begin{equation}
    \pi^*(\mathbf{a}\mid s)
    \propto
    \pi_\beta(\mathbf{a}\mid s)
    \exp\left(Q_\phi(s,\mathbf{a})/\lambda\right),
\end{equation}
where $\lambda$ is the temperature. The resulting policy update can be written as a local regression objective along trajectories of the reference flow:
\begin{equation}
\begin{aligned}
\label{eq:loss-qam-prelim}
f_\delta(s,\mathbf{a}^w,w)=&f_\theta(s,\mathbf{a}^w,w)-f_\beta(s,\mathbf{a}^w,w) \\
\mathcal{L}_{\mathrm{QAM}}(\theta)=&
\mathbb{E}
\left[
\int_0^1
\left\|
\frac{2f_\delta(s,\mathbf{a}^w,w)}{\sigma_w}
+
\sigma_w \tilde{g}_w
\right\|_2^2
\,\mathrm{d}w
\right],
\end{aligned}
\end{equation}
where $\sigma_w=\sqrt{2(1-w)w}$ and $\tilde{g}_w$ is the adjoint state with terminal condition
\begin{equation} \label{eq:qam-terminal-prelim}
    \tilde{g}_1
    =
    -\nabla_{\mathbf{a}}
    \left[
    Q_\phi(s,\mathbf{a}^1)/\lambda
    \right].
\end{equation}
LWD uses QAM~\citep{li2026q} as its policy-extraction mechanism,  using the critic learned form DIVL to form local regression targets for the flow policy.

\section{Learning while Deploying}
\label{sec:LWD}

\begin{figure*}
    \centering
    \includegraphics[width=\linewidth]{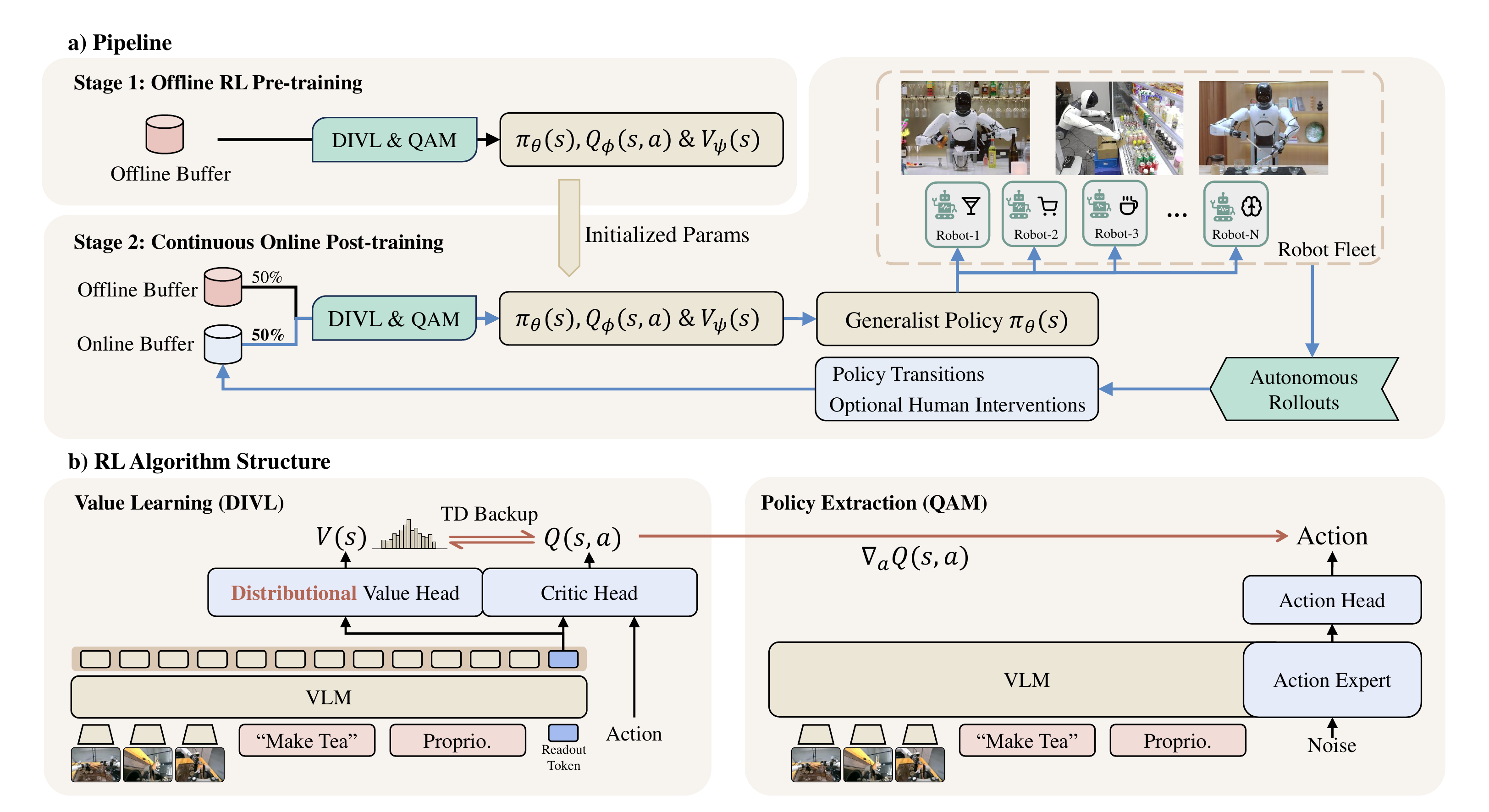}
    \caption{\textbf{LWD overview.} 
    (a) \textbf{Pipeline.} Training is organized into two stages. Stage 1 performs offline RL pre-training on an offline buffer. Stage 2 conducts continuous online post-training with mixed replay from both the static offline buffer and a continuously updated online buffer. A fleet of actors is autonomously deployed on diverse real-world robot tasks to collect online data and appends it to a continually updated online buffer.
    (b) \textbf{Algorithm structure.} A VLM-based model $\pi_{\theta}(s)$ maps states to action chunks through a policy head, which is optimized with QAM loss for policy extraction. In parallel, a critic $Q_\phi(s,\mathbf{a})$ and a distributional value $V_\psi(s)$ are trained with TD losses for value learning.}
    \label{fig:method}
\end{figure*}

LWD follows the offline-to-online procedure shown in Fig.~\ref{fig:method}(a). The offline stage trains the policy, critic, and distributional value model on a static replay buffer $\mathcal{B}_{\mathrm{off}}$, providing the initialization for deployment. The online stage deploys the current policy to a fleet of robot actors for autonomous rollouts, which populate $\mathcal{B}_{\mathrm{on}}$ with policy transitions and optional human interventions. The learner updates $V_\psi$, $Q_\phi$, and $f_\theta$ on mixed replay from $\mathcal{B}_{\mathrm{off}}\cup\mathcal{B}_{\mathrm{on}}$, and periodically redeploys the updated policy. This forms a data flywheel: robot rollouts expand replay, mixed replay updates the policy, and refreshed checkpoints are redeployed to the fleet.

This procedure contains two key algorithmic components. First, \emph{Distributional Implicit Value Learning (DIVL)} trains the critic $Q_\phi$ and the distributional value model $V_\psi$ for value learning. Second, QAM-based policy extraction updates the flow policy $f_\theta$ using the action gradient of $Q_\phi$ learned from DIVL.

\subsection{Distributional Implicit Value Learning}
\label{subsec:DIVL}
Distributional Implicit Value Learning (DIVL) is the value-learning component of LWD. It learns a distribution over replay action-values and uses a quantile of this distribution as the bootstrap target for the chunk-level critic $Q_\phi(s_t,\mathbf{a}_t)$. This design keeps the asymmetric bootstrap principle of IQL~\citep{kostrikov2021offline} while avoiding a single scalar expectile target.

Concretely, the distributional value model $V_\psi(s_t)$ represents the state-conditioned distribution of dataset action-values~~\citep{bellemare2017distributional}:
\begin{equation} \label{eq:dist_value}
    p_\psi(v \mid s_t) = P\!\left(v = Q_{\phi}(s_t,\mathbf{a}_{t}) \mid \mathbf{a}_{t} \sim \mathcal{D}(\cdot\mid s_t)\right).
\end{equation}
where $\mathcal{D}(\cdot\mid s_t)$ denotes the empirical replay action distribution conditioned on $s_t$. Thus, $V_\psi(s_t)$ is not a scalar value estimate. Instead, it represents the distribution of scalar critic values assigned to replay actions at state $s_t$.

We fit this distribution by minimizing the negative log-likelihood of scalar critic targets from the exponential-moving-average (EMA) critic $Q_{\bar\phi}$:
\begin{equation}\label{eq:divl_value_ce}
\mathcal{L}_V(\psi) = \mathbb{E}_{(s_t,\mathbf{a}_{t})\sim\mathcal{D}} \Big[-\log p_\psi\!\big(Q_{\bar\phi}(s_t, \mathbf{a}_{t}) \mid s_t\big)\Big].
\end{equation}
In our implementation, $p_\psi$ is represented as categorical discretization; Appendix~\ref{appen-sec:divl-details} gives the details.

Compared with scalar regression used in IQL, such distributional parameterization is better matched to LWD. Prior work~\citep{kumar2022offline} finds that a categorical distributional representation of return values is helpful in diverse multi-task offline RL settings. Moreover, it supports two designs used below: the bootstrap statistic is selected as a quantile of $p_\psi(v\mid s_t)$ without refitting a scalar value function, and the entropy of $p_\psi(v\mid s_t)$ provides the uncertainty signal for adapting $\tau$.

We use the $\tau$-quantile of $V_\psi(s_t)$ as the bootstrap statistic:
\begin{equation}
\mathrm{Quant}_\tau\!\big(V_\psi(s_t)\big)
\triangleq
\inf\left\{v : F_\psi(v\mid s_t) \ge \tau\right\}.
\label{eq:divl_quantile}
\end{equation}
where $F_\psi(v\mid s_t)$ be the cumulative distribution function induced by $p_\psi(v\mid s_t)$. 
This yields the TD target
\begin{equation}
y_Q = \mathbf{r}_{t} + \gamma^{H} \, \mathrm{Quant}_\tau\!\big(V_{\psi}(s_{t+H})\big),
\label{eq:divl_td_target}
\end{equation}
and the critic loss
\begin{equation}
\mathcal{L}_Q(\phi) = \mathbb{E}_{(s_t,\mathbf{a}_{t},\mathbf{r}_{t},s_{t+H})\sim\mathcal{D}} \Big[\big(Q_\phi(s_t, \mathbf{a}_{t}) - y_Q\big)^2\Big].
\label{eq:divl_q_loss}
\end{equation}

The $\tau$-quantile is an in-distribution optimistic bootstrap statistic over replay actions, rather than an explicit max backup over the full action space. This fits the offline RL setting, where the target should favor high-value replay actions without extrapolating aggressively beyond the data. IQL addresses the same issue with scalar expectile value regression. DIVL keeps this asymmetric value-learning principle, but realizes it through a distributional model and a quantile statistic.

To make this connection explicit, we write the value target under a generalized asymmetric loss family:
\begin{equation}
\rho_{\tau, p}(u) = |\tau - \mathbb{I}(u < 0)| \cdot |u|^p,
\end{equation}
where $p=2$ gives the expectile form used by IQL and $p=1$ gives the quantile form used by DIVL.

\medskip \noindent
\refstepcounter{proposition}\label{prop:distributional_view}%
\textit{Proposition \theproposition~(Distributional view of asymmetric value learning):} For any fixed asymmetric loss in this family, direct scalar regression and our two-step procedure of fitting the value distribution and extracting the corresponding asymmetric statistic yield the same optimal scalar value.
\medskip

See Appendix~\ref{appen-sec:divl-proof} for proof. The proposition shows that DIVL's two-step procedure of distributional value estimation and $\tau$-quantile extraction has the same optimum as the corresponding direct asymmetric value regression objective.

This result supports using a quantile of the learned value distribution as the bootstrap target for a fixed $\tau$. The value of $\tau$ controls the optimism of this target: larger values select higher quantiles and give more optimistic targets, while smaller values give more conservative ones. In mixed-task replay, the same level of optimism is not appropriate for every state, so we adapt $\tau$ using uncertainty in the learned value distribution.

Specifically, we use the normalized entropy of the categorical distribution $p_\psi(\cdot\mid s_{t+H})$ as the uncertainty signal:
\begin{equation}
\mathcal{H}(s_{t+H})
=
-\frac{1}{\log C}
\sum_{c=1}^{C}
p_{\psi,c}(s_{t+H})\log p_{\psi,c}(s_{t+H}),
\label{eq:dynamic_tau_entropy}
\end{equation}
where $C$ is the number of categories and $p_{\psi,c}(s_{t+H})$ is the probability assigned to category $c$.
Then, the adaptive schedule is
\begin{equation}
\tau(s_{t+H}) = \mathrm{clip}\!\big(\tau_{\mathrm{base}} - \alpha\,\mathcal{H}(s_{t+H}),\; \tau_{\min},\; \tau_{\max}\big),
\label{eq:dynamic_tau}
\end{equation}
where $\tau_{\mathrm{base}}$ is the target for confident states, $\alpha \ge 0$ controls uncertainty sensitivity, and the hyperparameter values are reported in Appendix~\ref{appen-sec:training-details}. Diffuse distributions receive lower $\tau$ values to reduce overestimation, while concentrated distributions retain more optimistic targets. We treat $\tau(s_{t+H})$ as stop-gradient when computing the TD target.

\subsection{Policy Extraction via QAM}
\label{subsec:AM}
Policy extraction in LWD starts from a pretrained flow-matching VLA and aims to improve its action distribution using the DIVL critic. Existing offline RL methods often extract a policy without differentiating through $Q_\phi$, for example by advantage-weighted regression on replay actions~\citep{peng2019advantage,nair2020awac,kostrikov2021offline,zhang2025energyweighted}. This update is poorly matched to flow-based VLA policies, since it requires evaluating the log likelihood of action chunks under the multi-step denoising process of the flow policy. More generally, the KL-regularized policy improvement target has a Boltzmann form, whose normalizer requires integrating over high-dimensional action chunks.

An alternative is to use the first-order action gradient $\nabla_{\mathbf{a}} Q_\phi(s,\mathbf{a})$ to improve sampled action chunks. For flow policies, however, applying this update via direct backpropagation through the full multi-step generation process is computationally expensive and numerically unstable (see Appendix~\ref{appen-sec:flow-backprop} for analysis). This makes direct critic backpropagation difficult to use as the optimization method for large VLA policies.

We therefore use QAM for policy extraction~\citep{li2026q} as shown in the right of Fig.~\ref{fig:method}(b). As outlined in Section~\ref{subsec:qam}, QAM reformulates trajectory-level policy optimization into a local regression objective along the reference flow. Specifically, the DIVL critic $Q_\phi$ supplies the reward-informed gradient to initialize the terminal adjoint state $\tilde{g}_1$, which in turn guides the refinement of the policy vector field (Eq.~\eqref{eq:qam-terminal-prelim}).

In particular, we keep $f_\beta$ fixed as the behavior-cloned flow initialized before offline RL, and optimize $f_\theta$ throughout both offline and online training. For each replay minibatch, as shown in lines 5--7 of Algorithm~\ref{alg:learner_step}, we sample states and Gaussian noise, roll out $f_\beta$ to generate reference flow trajectories, evaluate $\nabla_{\mathbf{a}} Q_\phi(s,\mathbf{a})$ at the generated endpoint, solve the adjoint dynamics, and regress $f_\theta$ toward the resulting local targets.

\begin{algorithm}[t]
  \caption{LWD: Offline-to-Online Training Pipeline}
  \label{alg:lwd_pipeline}
  \begin{algorithmic}[1]
  \Require
  offline buffer $\mathcal{B}_{\mathrm{off}}$; demonstration dataset $\mathcal{D}_{\mathrm{demo}}\subset\mathcal{B}_{\mathrm{off}}$;
  online buffer $\mathcal{B}_{\mathrm{on}}$;
  robot actor fleet $\mathcal{F}$;
  offline budget $N_{\mathrm{off}}$; online budget $N_{\mathrm{on}}$; actor-sync period $N_{\mathrm{sync}}$.
  \State Pretrain policy $\pi_\theta \leftarrow \mathcal{D}_{\mathrm{demo}}$
  \State Set fixed reference policy $\pi_\beta\leftarrow \pi_\theta$
  \State Initialize $Q_\phi$, $V_\psi$; set target $Q_{\bar\phi}\leftarrow Q_\phi$

\vspace{4pt}
\Statex \texttt{// Stage 1: Offline Pretraining}
  \For{$i \leftarrow 1:N_{\mathrm{off}}$}
    \State Sample mini-batch $\mathcal{B}^{\text{mini}}\sim\mathcal{B}_{\mathrm{off}}$
    \State {\small $(Q_\phi, V_\psi, \pi_\theta, Q_{\bar\phi}) \leftarrow \textsc{Learner}(\mathcal{B}^{\text{mini}};\, Q_\phi, V_\psi, \pi_\theta, \pi_\beta, Q_{\bar\phi})$}
  \EndFor

\vspace{4pt}
\Statex \texttt{// Stage 2: Continuous Online Training}

\vspace{2pt}
\Statex \textbf{Robot actor process (Asynchronously):}
   \State Deploy $\pi_\theta$ to each robot from $\mathcal{F}$
  \While{online training is active}
    \State $\mathrm{done} \leftarrow False$; $T \leftarrow 0$
    \While{not $\mathrm{done}$}
      \State Execute $\mathbf{a} \leftarrow \pi_\theta(s)$ until $\mathrm{done}$
      \If{intervention is required}
        \State Human intervents $\mathbf{a} \leftarrow \mathbf{a}_{H}$
      \EndIf
      \State $s' \leftarrow$ \textit{UpdateObs(s, $\mathbf{a}$)}
      \State $\mathrm{done} \leftarrow \mathbb{I}[TimeLimit \lor Failure \lor Success]$
      \State $r \leftarrow \mathbb{I}[\mathrm{done} \land Success]$
      \State $T \leftarrow T+1$
    \EndWhile
    \State $\mathbf{r} \leftarrow \textit{UpdateChunkedReward}(r)$
    \State $\mathcal{B}_{\mathrm{on}} \leftarrow \mathcal{B}_{\mathrm{on}} \cup \{(s_t,\mathbf{a}_t,\mathbf{r}_t, s'_{t+H})\}$
    \State $\pi_\theta \leftarrow \textit{FetchNewPolicy}(\pi_\theta^{new})$
  \EndWhile

\vspace{2pt}
\Statex \textbf{Central learner process (Asynchronously):}
  \For{$j \leftarrow 1:N_{\mathrm{on}}$}
    \State Sample mini-batch $\mathcal{B}^{\text{mini}}\sim \{\mathcal{B}_{\mathrm{off}}\cup\mathcal{B}_{\mathrm{on}}\}$
    \State {\small $(Q_\phi, V_\psi, \pi_\theta, Q_{\bar\phi}) \leftarrow \textsc{Learner}(\mathcal{B}^{\text{mini}};\, Q_\phi, V_\psi, \pi_\theta, \pi_\beta, Q_{\bar\phi})$}
    \If{$j \bmod N_{\mathrm{sync}} = 0$}
      \State Deploy latest policy $\pi_\theta$ to each robot from $\mathcal{F}$
    \EndIf
  \EndFor

  \State \Return $Q_\phi$, $V_\psi$, $\pi_\theta$
  \end{algorithmic}
\end{algorithm}

\begin{algorithm}[t]
  \caption{$\textsc{Learner}$: Single Update of DIVL and QAM}
  \label{alg:learner_step}
  \begin{algorithmic}[1]
  \Require
  mini-batch $\mathcal{B}^{\text{mini}}=\{(s_t,\mathbf{a}_{t},\mathbf{r}_{t},s_{t+H})\}$;
  critic $Q_{\phi}$ with target $Q_{\bar\phi}$; distributional value $V_{\psi}$;
  policy $\pi_{\theta}$ with reference policy $\pi_{\beta}$;
  EMA rate $\rho$.
    \Statex \texttt{// Distributional Implicit Value Learning}
    \State Update $\psi$ by minimizing Eq.~\eqref{eq:divl_value_ce}
    \State Compute TD target $y_Q$ via Eq.~\eqref{eq:n-step_TD_target}
    \State Update $\phi$ by minimizing Eq.~\eqref{eq:divl_q_loss}
    \State $\bar\phi \leftarrow \rho\,\bar\phi + (1-\rho)\,\phi$
    \Statex \texttt{// Policy Extraction via QAM}
    \State Sample Gaussian noise $\mathbf{a}_{t}^0\sim\mathcal{N}(0,I)$
    \State Roll out the reference trajectory $\{\mathbf{a}_t^w\}_{w\in[0,1]}$ via $\pi_\beta$
    \State Set the endpoint $\mathbf a_t^1 = \mathbf a_t$
    \State Update $\theta$ by minimizing Eq.~\eqref{eq:loss-qam-prelim} with $\tilde g_1$ set from action gradient $\nabla_{\mathbf{a}}Q_\phi(s, \mathbf{a}_t^1)$ via Eq.~\eqref{eq:qam-terminal-prelim}

    \State \Return $(Q_\phi, V_\psi, \pi_\theta, Q_{\bar\phi})$
  \end{algorithmic}
\end{algorithm}

\subsection{Offline to Online RL Training Pipeline}
\label{subsec:offline-to-online-pipeline}
Following the LWD loop in Fig.~\ref{fig:method}(a) as introduced in the opening of Section~\ref{sec:LWD}, post-training proceeds in two stages that share the same value-learning and policy-extraction objectives but differ in data source.

The offline stage trains on an offline buffer $\mathcal{B}_{\mathrm{off}}$ as shown in the Stage 1 of Fig.~\ref{fig:method}(a) and lines 4--7 of Algorithm~\ref{alg:lwd_pipeline}. This offline buffer contains three sources: \textit{demonstrations}, expert-collected successful trajectories; \textit{rollouts}, generated by historical policies, including both successes and failures; and \textit{play data}, consisting of human-guided exploration of failure modes. All three sources are converted into the same chunked transition format as online replay, with terminal success or failure labels used to assign sparse binary rewards. The details of data structure are shown in Table~\ref{tab:data-composition}. LWD applies the offline buffer to pre-train the policy $\pi_\theta$, critic $Q_\phi$ and distributional value $V_\psi$, providing a strong initialization for deployment and training in the online stage. 

Moreover, since long-horizon tasks last thousands of steps and have extremely sparse rewards, the one-step target in Eq.~\eqref{eq:divl_td_target} can propagate success signals slowly. We therefore use an $n$-step chunk-level TD target in the offline stage to cold-start the critic and distributional value model:
\begin{equation} \label{eq:n-step_TD_target}
    y_Q = \sum_{i=0}^{n-1} \gamma^{iH}\mathbf{r}_{t+iH} + \gamma^{nH}\mathrm{Quant}_{\tau(s_{t+nH})}\big(V_\psi(s_{t+nH})\big),
\end{equation}
where $n=1$ for short tasks such as grocery restocking tasks and $n=10$ for long-horizon tasks. If an episode terminates within the $n$-step window, we truncate the return at the terminal chunk and remove the bootstrap term. This target accelerates sparse reward propagation through the fixed offline replay buffer. During online training, we found long multi-step targets less effective. One possible reason is that online trajectories mix policy transitions with human interventions. Longer backups are more likely to cross these sources, so the TD path may not correspond to a single policy execution. Since the critic and value model already have offline initialization, we therefore use 1-step chunk-level TD targets for online updates.

The online stage deploys the offline-initialized policy to the robot fleet, as shown in Stage 2 of Fig.~\ref{fig:method}(a) and lines 8--31 of Algorithm~\ref{alg:lwd_pipeline}. Robots execute the current policy checkpoint and asynchronously stream \textit{policy transitions} into an online buffer $\mathcal{B}_{\mathrm{on}}$. When a rollout requires correction, as judged by human, the operator may intervene. Intervention segments are stored in $\mathcal{B}_{\mathrm{on}}$ as regular online replay transitions with the executed corrective actions, and rewards are assigned using the same terminal success or failure labels as autonomous rollouts. Thus, online replay contains both autonomous policy transitions and human-intervention transitions~\citep{hilserl}. Online Training continues with the same value-learning and policy-extraction objectives on mixed replay from $\mathcal{B}_{\mathrm{off}} \cup \mathcal{B}_{\mathrm{on}}$, while updated policy checkpoints are periodically published back to the robots. 

\subsection{Architectures}
\label{subsec:network-architecture}
Fig.~\ref{fig:method}(b) shows the concrete neural network architecture used by LWD. The policy and value/critic networks are separate modules, isolating action generation from value and critic optimization. Only the policy checkpoint is asynchronously distributed to the robot fleet for inference, while the value and critic networks remain on the centralized learner.

We implement \(V_\psi\) and \(Q_\phi\) with a shared Gemma3--SigLIP VLM backbone and separate prediction heads. The Gemma~3 language module and SigLIP vision encoder are initialized from publicly released Gemma~3-270M-IT ~\cite{gemm3techincalreport} and SigLIP-So400M checkpoints ~\cite{zhai2023sigmoid}, while the visual projection layer and value/critic heads are initialized from scratch. 

Following the use of readout tokens as compact transformer representations~\citep{dosovitskiy2020image,ranftl2021vision,li2023blip}, we apply the shared backbone to the multimodal sequence for state $s_t$ and denote the final hidden state of the readout token by $z_t$, which serves as the state representation for both value and critic prediction. The value head 
predicts logits over a fixed categorical support.
Following the C51 projection~\citep{bellemare2017distributional}, the scalar supervision target $Q_{\bar{\phi}}(s_t,\mathbf{a}_t)$ is clipped to the value support and linearly projected onto its two neighboring atoms, yielding a target distribution $m_t$. 

The critic conditions on both the state representation \(z_t\) and the action chunk \(\mathbf{a}_t\). The action chunk is encoded with a learned temporal attention pooling layer and concatenated with \(z_t\). The resulting representation is fed into two scalar critic heads in a clipped double-Q design, where the minimum critic estimate is used for DIVL target construction and TD backups to mitigate overestimation.

The actor follows the \(\pi_{0.5}\) flow-based VLA architecture~\citep{pi05}. It consists of a PaliGemma vision-language backbone, instantiated with a Gemma-2B language model and a SigLIP vision encoder, together with a Gemma-300M action expert for flow-based action generation.

In the offline RL stage, both the actor and the value/critic networks are fully fine-tuned; the resulting weights initialize online training. During online QAM updates, the policy VLM backbone is frozen and only the action expert is updated, while the value and critic networks continue to be fully fine-tuned on mixed replay. This design keeps online policy updates efficient and preserves the pretrained vision-language representations, while allowing the value and critic networks to adapt to the evolving replay distribution and provide updated policy-improvement signals.

\section{Experimental Evaluations}
\label{sec:exp}
We evaluate LWD on eight real-world manipulation tasks including grocery stocking and long-horizon manipulation tasks such as tea making, juice making and more. We compare against the reference policy, SFT~\citep{lipman2022flow} and two representative post-training baselines, RECAP~\citep{pi06} and HG-DAgger~\citep{kelly2019hg}.
Our experiments seek answers to whether deployment-time online updates from a shared robot fleet improve over static or offline policies in the same task setting; how LWD compares with the baselines; whether the learned value function provides a useful progress signal under sparse terminal rewards, and which design choice of DIVL contributes to the observed gains.
The main results addresses method performance, the value-function visualization diagnoses whether sparse-reward value estimates track task progress, and the ablations isolate the DIVL value-estimation design choices.

\begin{figure*}[htbp]
    \centering
    \includegraphics[width=0.95\linewidth]{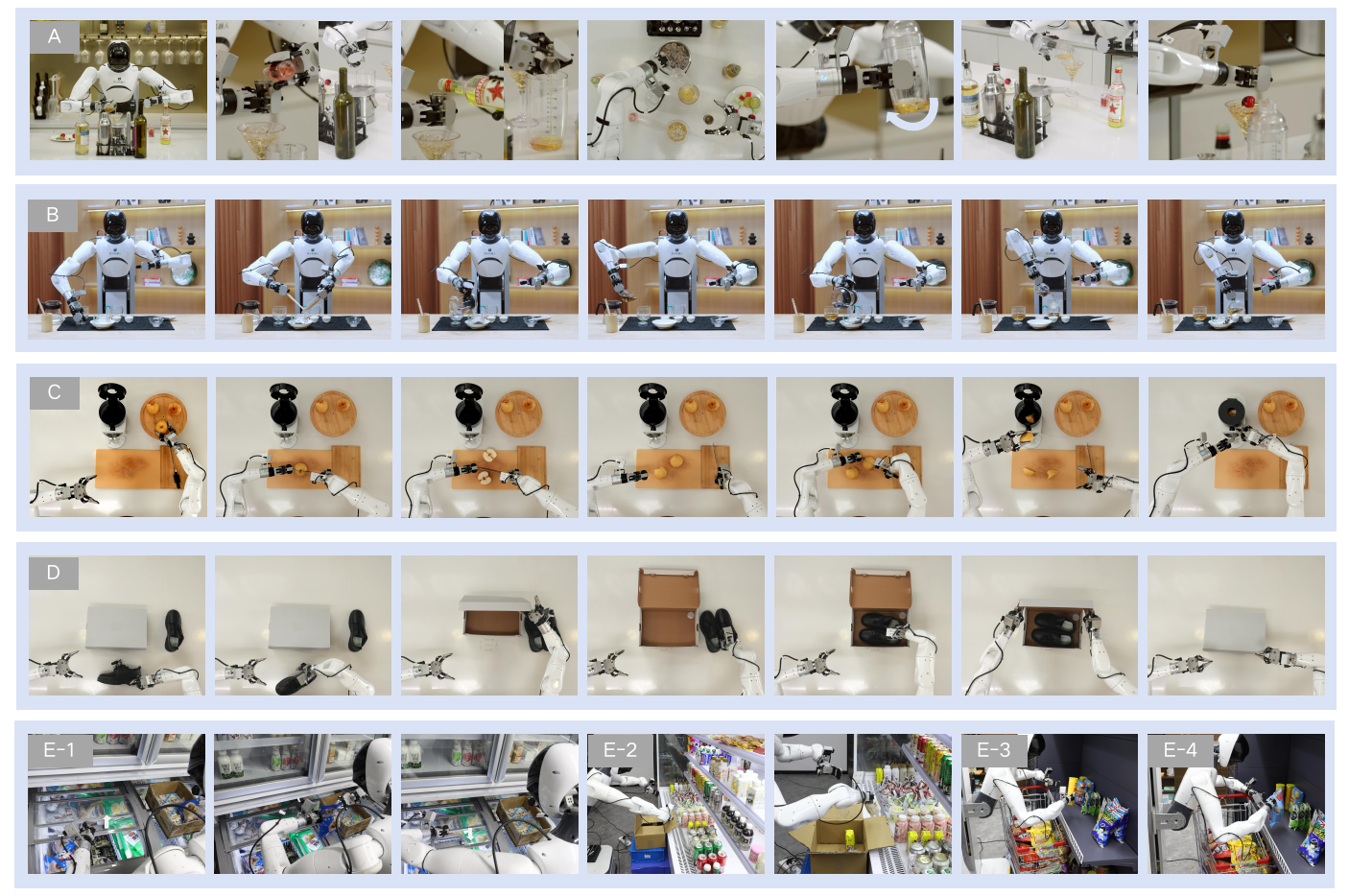}
    \caption{\textbf{Illustrations of our evaluation tasks.} Panels A--D show the four long-horizon tasks, and Panel E summarizes the four grocery restocking tasks.
    \textbf{(A) Make Cocktail:} A sequence of robot manipulation actions for cocktail making: measuring and mixing multiple liquors in a shaker, adding ice, shaking the cocktail, pouring it into a stemmed glass, and garnishing it with a cherry.
    \textbf{(B) Brew Gongfu Tea:} A robot manipulation sequence for Gongfu tea preparation: adding tea leaves, rinsing and draining, brewing with hot water, transferring the tea to a fairness pitcher, distributing it into three teacups, and serving.
    \textbf{(C) Make Fruit Juice:} The sequence for fruit juicing, including cutting and reorienting the fruit, slicing it into pieces, transferring the pieces into a juicer, closing the lid, and rotating the control knob to start juicing.
    \textbf{(D) Pack Shoes:} A manipulation sequence of packing shoes into a shoebox and placing the shoebox neatly.
    \textbf{(E) Grocery Restocking Tasks:} Robot manipulation tasks in various grocery scenarios, including freezer restocking involving door manipulation, open-cooler restocking with carton handling, and flat-shelf restocking with misplacement correction.
    Together, the suite stresses semantic grounding, contact-rich manipulation, long-horizon execution, and recovery from execution errors.
    }
    \label{fig:exp_task_frame}
\end{figure*}

\subsection{Experimental Setup}

\subsubsection{Tasks, Evaluation, and Robots}
\paragraph{Tasks} We evaluate LWD on eight real-world tasks, as shown in Fig.~\ref{fig:exp_task_frame}. The grocery restocking tasks consist of four distinct tasks: flat-shelf restocking, misplaced-item correction, freezer restocking with door operation, and open-cooler restocking with carton handling. Together, they test the policy's ability to follow language instructions and generalize semantically across realistic store scenarios.
In each task, the robot must identify the object specified by the instruction among cluttered candidates, handle variations in shelf layout and container geometry, and complete the required placement.
The evaluation varies object instances, clutter, shelf and container layouts, language instructions, and store configurations.

We also evaluate our methods on four long-horizon tasks: brewing Gongfu Tea, making Fruit Juice, making Cocktail, and packing shoes into a Shoebox. Each episode typically lasts 3--5 minutes and contains 5--8 annotated subtasks, creating long-range dependencies across planning, manipulation, and recovery. Success requires stable multi-stage executions with precise contact-rich skills, including grasp adjustment, container handling, pouring, tool use, and final placement. Evaluation episodes include natural reset variability in object poses, tool locations, ingredients, scene initialization, perturbations, and occasional retry or recovery situations. 

\paragraph{Evaluation metrics} We report task-level scores for all tasks, with different scoring protocols for the two task groups. For the grocery restocking tasks, we follow the protocol of SOP~\citep{pan2026sop}: an episode is successful if the robot follows the right language instruction and task completion within the time limit, yielding a binary success rate. For long-horizon tasks, we report a step-wise success score. Each annotated sub-step is scored as 1 (fully autonomous success), 0.5 (success with minor imperfection, or success with a single retrial), or 0 (failure after multiple attempts), and the task score is the average across sub-steps. The score is assigned by trained human evaluators according to a predefined rubric that is applied consistently across methods and tasks. 
We additionally report cycle time on long-horizon tasks to evaluate execution efficiency. Cycle time is computed over both successful and failed attempts, with failed trajectories clipped at predefined task-specific timeout thresholds.

\begin{figure}[tbp]
    \includegraphics[width=\linewidth]{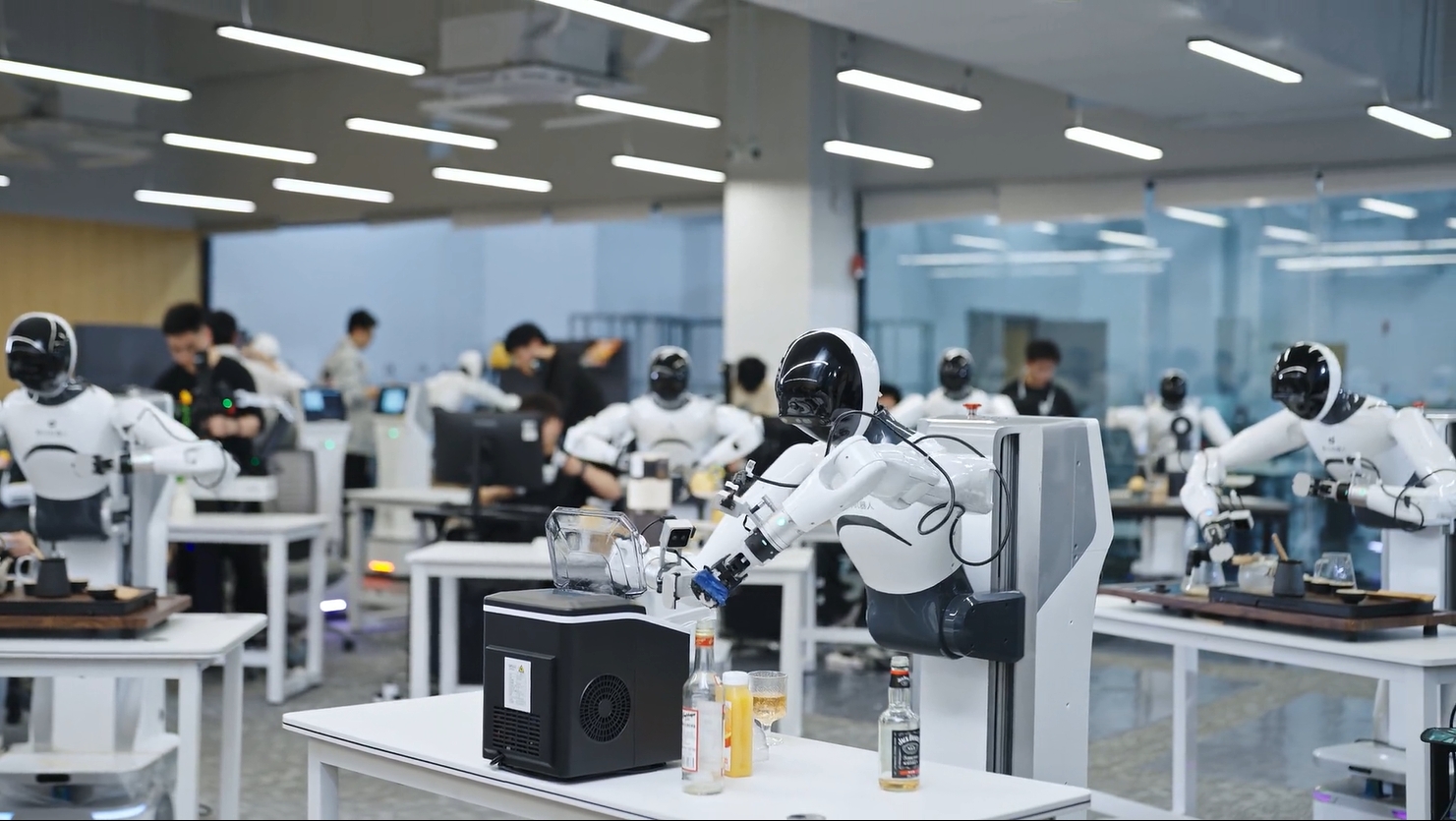}
    \captionof{figure}{
    \textbf{Fleet of robots.} LWD performs online training across a fleet of 16 robots, continually improving a single generalist policy on multiple tasks.}
    \label{fig:fleet}
\end{figure}
\paragraph{Robot fleet setup} All experiments are conducted on the Agibot G1 dual-arm manipulation platform. Each G1 robot has two 7-DoF arms with parallel-jaw grippers and three RGB cameras (one head-view and two wrist-view). The policy runs joint-position control at 30\,Hz. 
As shown in Fig.~\ref{fig:fleet}, we deploy a fleet of 16 robots for concurrent rollout collection during online training: 4 robots for the grocery restocking tasks and 3 robots for each long-horizon task. 
The fleet is connected to a distributed actor-learner system: edge actors upload complete episodes, a centralized learner fetches versioned replay data, and publishes the updated policies to each actor; more details can be seen in Appendix~\ref{appen-sec:lwd-infra}. 
For each online experiment, each method is allocated a 4-hour wall-clock budget, corresponding to approximately 60 total hours of online data collected across the robot fleet. 
Robots collect rollouts asynchronously, and episodes from all tasks are pooled into a single online replay buffer for updating the shared policy. The buffer contains both autonomous rollouts and human intervention segments when intervention is required. The learner broadcasts the updated shared policy to the robot fleet every 50 training steps.
Additional training details and hyperparameters are provided in Appendix~\ref{appen-sec:training-details}.

\subsubsection{Baselines and Reference Policies}
We compare against two post-training baselines, RECAP and HG-DAgger, and SFT as a reference policy. 
SFT only utilizes human demonstrations with standard flow-matching loss.
RECAP~\citep{pi06} starts from the reference policy and performs iterative post-training on autonomous rollouts. We preserve its advantage-conditioned policy-improvement recipe, but implement it in a multi-task setting. 
HG-DAgger~\citep{kelly2019hg} also starts from the reference policy, then uses online successful rollouts for training. 
Implementation details and hyperparameters are provided in Appendix~\ref{appen-sec:experiment-details}.

\begin{figure*}
    \centering
    \includegraphics[width=0.98\linewidth]{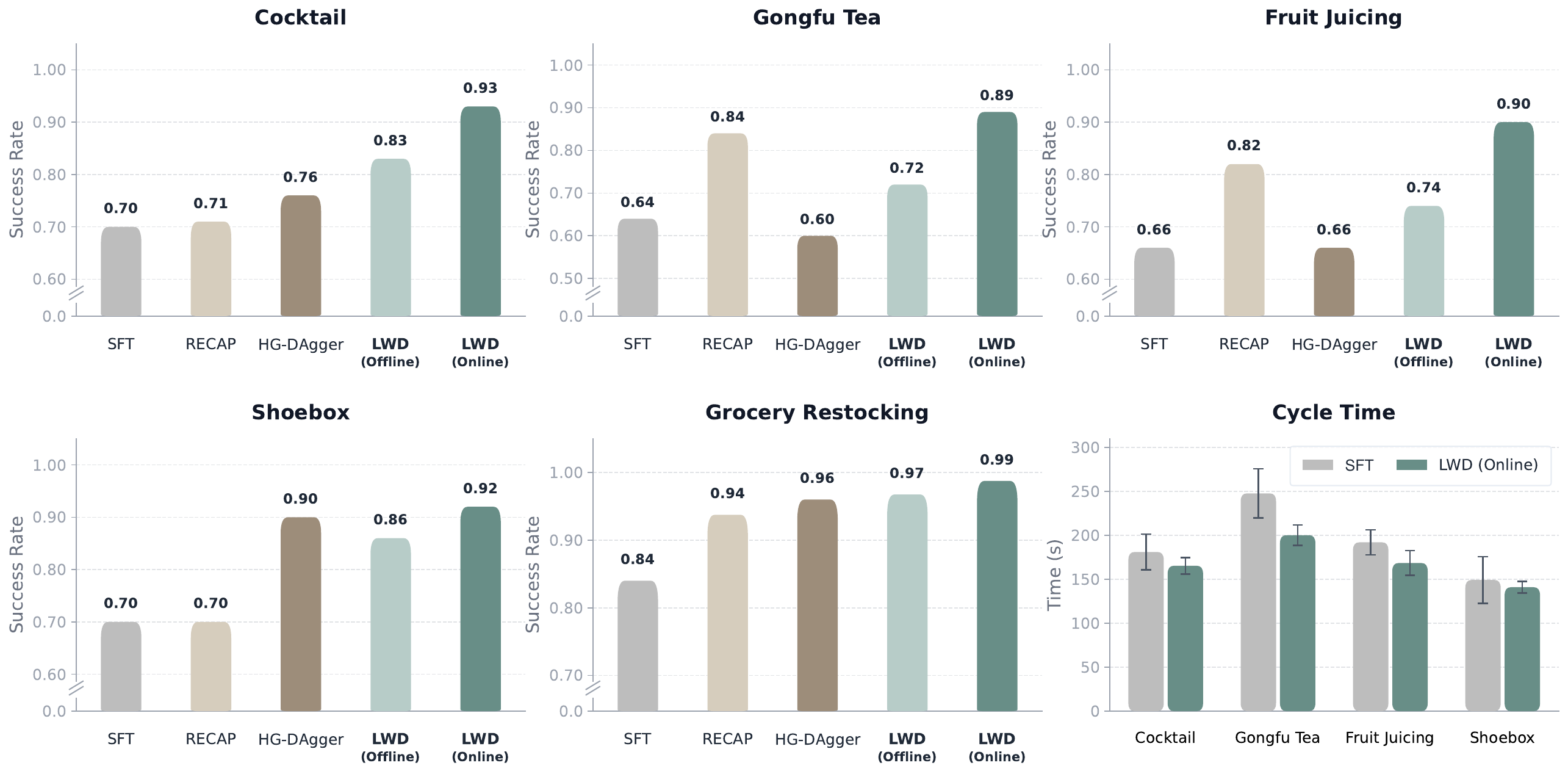}
    \caption{\textbf{Success scores and cycle-time comparison.}
    LWD achieves higher success scores while reducing mean cycle time relative to the static SFT reference policy. Complete results are shown in Table~\ref{tab:main-results}. 
    }
    \label{fig:exp_res_success_rate}
\end{figure*}

\subsection{Main Results}
Table~\ref{tab:main-results} reports the main quantitative results across all eight real-world tasks. LWD (Online) achieves an average score of $0.95$, outperforming all baselines across the evaluated tasks and maintaining strong performance on both short-horizon and long-horizon tasks.

\begin{table*}[t]
\centering
\caption{\textbf{Complete results on eight real-world manipulation tasks}, covering four grocery restocking tasks and four long-horizon tasks. We report task success rate for each task (binary success for grocery restocking tasks; average step-wise score across sub-steps for long-horizon Tasks) and the average across all eight tasks in the last column. The best result per column is shown in bold. Our LWD (Online) attains the best overall average (0.95) and achieves the top score on all four long-horizon tasks, while remaining at or near the best on the grocery restocking tasks.}
\label{tab:main-results}
\setlength{\tabcolsep}{4pt}
\small
\begin{tabular}{lccccccccc}
\toprule
\multirow{2}{*}{\textbf{Method}} & \multicolumn{4}{c}{\textbf{Grocery Restocking Tasks}} & \multicolumn{4}{c}{\textbf{Long-Horizon Tasks}} & \multirow{2}{*}{\textbf{Average}} \\
\cmidrule(lr){2-5}\cmidrule(lr){6-9}
& \textbf{Restocking} & \textbf{Correction} & \textbf{Freezer} & \textbf{Open-Cooler} & \textbf{Gongfu Tea} & \textbf{Fruit Juice} & \textbf{Cocktail} & \textbf{Shoebox} & \\
\midrule
SFT~\citep{lipman2022flow} & 0.70 & 0.88 & 0.83 & 0.95 & 0.64 & 0.66 & 0.70 & 0.70 & 0.76 \\
RECAP \cite{pi06} & 0.95 & 0.96 & 0.94 & 0.95 & 0.84 & 0.82 & 0.71 & 0.70 & 0.85 \\
HG-DAgger \cite{kelly2019hg} & \textbf{1.00} & 0.92 & 0.92 & \textbf{1.00} & 0.60 & 0.66 & 0.76 & 0.90 & 0.85 \\
LWD (Offline, Ours) & \textbf{1.00} & \textbf{1.00} & 0.92 & 0.95  & 0.72 & 0.74 & 0.83 & 0.86 & 0.88 \\
\textbf{LWD (Online, Ours)} & \textbf{1.00} & \textbf{1.00} & \textbf{0.97} & 0.98 & \textbf{0.89} & \textbf{0.90} & \textbf{0.93} & \textbf{0.92} & \textbf{0.95} \\
\bottomrule
\end{tabular}
\end{table*}

The benefit of LWD is more pronounced on long-horizon tasks. LWD (Online) reaches an average long-horizon step-wise score of $0.91$, outperforming SFT ($0.68$), RECAP ($0.77$), HG-DAgger ($0.73$), and LWD (Offline) ($0.79$). This improvement can be attributed to a consistent offline-to-online RL training pipeline and more complete utilization of available data. LWD (Online) incorporates successful demonstrations, play data, and both successful and failed online trajectories into reward-based policy improvement, enabling the policy to continuously identify and mitigate failure modes encountered during deployment.

LWD (Offline) builds on the reference policy and improves it through offline reinforcement learning. It trains on an offline replay buffer containing successful demonstrations, failed rollouts, and diverse play data, allowing the policy to exploit reward and outcome information beyond imitation-only supervision. We further observe that HG-DAgger yields only limited gains over the reference policy on long-horizon tasks and can even degrade performance on some tasks. A likely reason is that DAgger-style training relies on human correction data, whose variability can introduce inconsistencies and provide limited exploration of the broader state space. In contrast, RL can exploit a wider range of states and directly optimize task-specific rewards. For long-horizon tasks, terminal success signals can be propagated to earlier decision steps through TD backups, improving value estimation across different task stages and providing a stronger learning signal for policy improvement.

On the grocery restocking tasks, all methods except SFT achieve high scores, leaving limited room for improvement. Even in this saturated regime, LWD (Online) remains at or near the best-performing result on every grocery task. This indicates that LWD provides benefits beyond long-horizon tasks while preserving the generalist behavior of the shared policy during online learning.

Fig.~\ref{fig:exp_res_success_rate} further reports the mean and standard error of cycle time on long-horizon tasks. LWD reduces mean cycle time by $23.75$ seconds compared with reference policy. 
This efficiency gain is consistent with the critic-guided policy update. The learned value function favors action chunks that make reliable task progress. As a result, the policy reduces hesitations, retries, and unstable intermediate behaviors, rather than only improving eventual task completion.


\begin{figure*}
    \centering
    \includegraphics[width=1.0\linewidth]{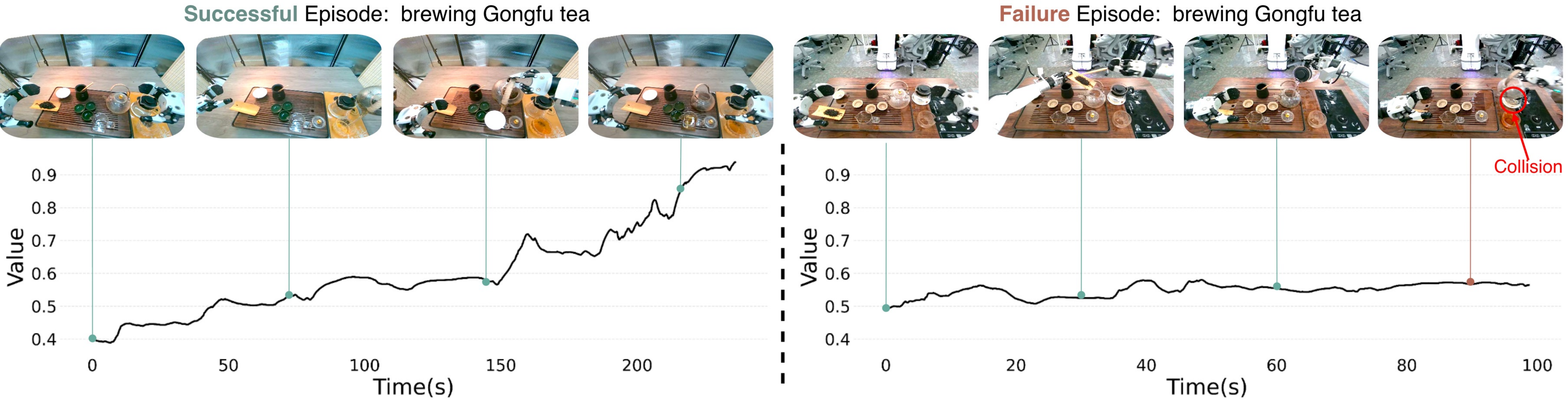}
    \caption{\textbf{Visualizations of value learning.} We plot quantile values of the learned distributional value function $V$ over time for representative Gongfu Tea episodes. The left trajectory succeeds and the right trajectory fails. The curves are qualitative diagnostics and are consistent with the learned value estimate tracking task-progress differences in these examples.
    }
    \label{fig:value_vis}
\end{figure*}

Fig.~\ref{fig:value_vis} visualizes the value estimate during a successful and a failed Gongfu Tea episode. In the successful episode, the value tends to increase as the robot completes key sub-steps and approaches task completion, suggesting that the learned value can reflect progress despite sparse terminal rewards. In the failure episode, the value fluctuates locally but remains lower after the execution stops making progress toward the annotated task milestones. Additional visualizations of the predicted value distributions are provided in Appendix Fig.~\ref{fig:value_dist_vis}.

\subsection{Ablation Study}

\subsubsection{Value Learning Design}
\begin{table}[t]
\centering
\caption{\textbf{Ablation of value learning design.} We report the average success rate on the short-horizon (grocery restocking tasks) and long-horizon tasks under offline and online settings.}
\label{tab:ablation-average-results}
\setlength{\tabcolsep}{1pt}
\scriptsize
\begin{tabular*}{\columnwidth}{@{\extracolsep{\fill}}lcccc@{\hspace{1em}}}
\toprule
\multirow{2}{*}{\textbf{Method}} &
\multicolumn{2}{c}{\textbf{Short-Horizon}} &
\multicolumn{2}{c}{\textbf{Long-Horizon}} \\
\cmidrule{2-3}\cmidrule{4-5}
& \textbf{Offline} & \textbf{Online} & \textbf{Offline} & \textbf{Online} \\
\midrule
Expectile Regression  & \makebox[4.8em][c]{\makebox[0pt][c]{0.96}} & \makebox[4.8em][c]{\makebox[0pt][c]{0.97}} & \makebox[4.8em][c]{\makebox[0pt][c]{0.72}} & \makebox[4.8em][c]{\makebox[0pt][c]{0.78}} \\
DIVL (Ours) & \makebox[4.8em][c]{\makebox[0pt][c]{0.97}\makebox[0pt][l]{\hspace{1em}\scalebox{0.9}{\tiny\textcolor{red}{(+1.0\%)}}}} & \makebox[4.8em][c]{\makebox[0pt][c]{0.99}\makebox[0pt][l]{\hspace{1em}\scalebox{0.9}{\tiny\textcolor{red}{(+2.1\%)}}}} & \makebox[4.8em][c]{\makebox[0pt][c]{0.79}\makebox[0pt][l]{\hspace{1em}\scalebox{0.9}{\tiny\textcolor{red}{(+9.7\%)}}}} & \makebox[4.8em][c]{\makebox[0pt][c]{0.91}\makebox[0pt][l]{\hspace{1em}\scalebox{0.9}{\tiny\textcolor{red}{(+\textbf{16.7}\%)}}}} \\
\bottomrule
\end{tabular*}
\end{table}

We compare DIVL with scalar expectile value regression while keeping all other components fixed. DIVL outperforms the scalar baseline on all tasks, with larger gains on long-horizon tasks ($9.7\%$ in the offline stage and $16.7\%$ in the online stage). In fleet deployment, the replay buffer contains diverse successful, failed, and intervention trajectories collected across tasks and scenes. By compressing heterogeneous outcomes into a single expected value, a scalar value function blurs rare but reproducible high-return behaviors. A distributional value instead retains the return distribution, preserving these high-return modes and providing a more informative signal for policy improvement. Complete per-task results are reported in Appendix Table~\ref{tab:ablation_value_estimation}.

\subsubsection{Adaptive $\tau$ Strategy}
\begin{table*}[t]
\centering
\caption{\textbf{Ablation of the adaptive $\tau$ strategy in offline LWD.} We compare the adaptive $\tau$ schedule with a constant $\tau$ baseline. For the constant baseline, $\tau$ is set to the empirical average value of the adaptive schedule from the adaptive-$\tau$ run ($\tau=0.52$), while all other training components are kept unchanged. }
\label{tab:ablation_adaptive_tau}
\setlength{\tabcolsep}{4pt}
\small
\begin{tabular}{lccccccccc}
\toprule
\multirow{2}{*}{\textbf{Method}} & \multicolumn{4}{c}{\textbf{Grocery Restocking Tasks}} & \multicolumn{4}{c}{\textbf{Long-Horizon Tasks}} & \multirow{2}{*}{\textbf{Average}} \\
\cmidrule(lr){2-5}\cmidrule(lr){6-9}
& \textbf{Restocking} & \textbf{Correction} & \textbf{Freezer} & \textbf{Open-Cooler} & \textbf{Gongfu Tea} & \textbf{Fruit Juice} & \textbf{Cocktail} & \textbf{Shoebox} & \\
\midrule
LWD Offline, constant $\tau$ & 0.85 & 0.88 & \textbf{0.94} & \textbf{0.95} & 0.70 & \textbf{0.76} & 0.70 & \textbf{0.90} & 0.84 \\
LWD Offline, adaptive $\tau$ & \textbf{1.00} & \textbf{1.00} & 0.92 & \textbf{0.95} & \textbf{0.72} & 0.74 & \textbf{0.83} & 0.86 & \textbf{0.88} \\
\bottomrule
\end{tabular}
\end{table*}

We further ablate the adaptive $\tau$ strategy used in DIVL during offline LWD training. We compare the adaptive schedule against a constant-$\tau$ baseline, where the constant value ($\tau=0.52$) is set to the average $\tau$ observed from training statistics in the adaptive-$\tau$ run; all other components are kept identical.
Table~\ref{tab:ablation_adaptive_tau} shows that adaptive $\tau$ improves the average offline score from $0.84$ to $0.88$. Although the constant baseline is competitive on a few individual tasks, the adaptive schedule gives more consistent gains across all tasks, especially on Restocking, Correction, and Cocktail. This indicates that conditioning $\tau$ on distributional entropy helps calibrate bootstrap optimism, making targets more conservative under high uncertainty and more optimistic when the value estimate is confident.

\section{Conclusion}
\label{sec:conclusion}

We present Learning While Deploying (LWD), a large-scale real-world reinforcement learning framework for post-training generalist robot policies. 
LWD first initializes the policy from previously collected robot data, then continues improving it through online RL during deployment. The framework uses DIVL for value learning and QAM for policy extraction.
Across eight real-world manipulation tasks spanning grocery restocking and long-horizon manipulation, LWD delivers the best overall performance, with the most pronounced improvements on long-horizon tasks.

These results suggest a practical path toward large-scale real-world deployment of continuously improving robot systems. With LWD, deployment is not only the setting in which the policy is evaluated, but also the mechanism through which the policy improves. 
Interaction data collected from the robot fleet is aggregated into a shared learning process, enabling a generalist policy to continue improving across tasks. This is critical for real-world robotic systems that must operate in heterogeneous tasks and environments.

Our method has several limitations. First, the current online learning pipeline updates with a straightforward real-time schedule. This design may not be optimal for larger-scale deployment or long-term continual improvement. More efficient and stable update strategies remain an important direction for future work. Second, our long-horizon experiments rely on a single short language instruction for each task. However, complex tasks require stronger vision-language reasoning for task decomposition, as well as finer-grained prompts for closed-loop execution and error recovery. Third, our current policy learning framework does not explicitly model execution safety. Incorporating safety-aware learning and control mechanisms will be important for reliable real-world deployment. Despite these limitations, this work represents a step toward large-scale real-world deployment, with the long-term goal of continuously scaling robot learning systems for robust execution in unstructured environments.

\section{Acknowledgments}

We thank Qiyang Li for helpful discussions.

\bibliographystyle{IEEEtranN}
\bstctlcite{setting}
\bibliography{references}

\appendix

\subsection{Additional Method Details}

\subsubsection{Discretization of Distributional Value Model}
\label{appen-sec:divl-details}
We instantiate the distributional value model $V_\psi(s)$ with a fixed categorical support $\{V_i\}_{i=1}^{K}$ spanning $[v_{\min},v_{\max}]$. In our real-robot experiments, we use $K=201$ atoms over $[-0.1,1.1]$. The value head predicts logits over this support,
\begin{equation}
p_\psi(i\mid s) =
\mathrm{softmax}(V_\psi(s))_i,
\qquad i\in\{1,\ldots,K\}.
\end{equation}

For each replay sample $(s,\mathbf{a})$, the scalar target $Q_{\bar{\phi}}(s,\mathbf{a})$ is clipped to $[v_{\min},v_{\max}]$ and linearly projected onto the two neighboring atoms following the C51 projection~\citep{bellemare2017distributional}. This yields a target distribution $m(s,\mathbf{a})$ over atoms, and the distributional value model is trained by cross entropy:
\begin{equation}
\mathcal{L}_Z(\psi)
=
-\mathbb{E}_{(s,\mathbf{a})\sim\mathcal{D}}
\left[
\sum_{i=1}^{K}
m_i(s,\mathbf{a})\log p_\psi(i\mid s)
\right].
\end{equation}

The discrete CDF is
\begin{equation}
F_\psi(V_j\mid s)=\sum_{i\le j}p_\psi(i\mid s),
\end{equation}
and the quantile used in the DIVL TD target is obtained by selecting the first atom whose cumulative probability exceeds the desired level:
\begin{equation}
\mathrm{Quant}_{\tau}(V_\psi(s))
=
V_{\min\{j:F_\psi(V_j\mid s)\ge \tau\}}.
\end{equation}
The normalized entropy used in the adaptive $\tau$ strategy is
\begin{equation}
\mathcal{H}(s)
=
-\frac{1}{\log K}
\sum_{i=1}^{K}
p_\psi(i\mid s)\log p_\psi(i\mid s)
\in[0,1].
\end{equation}

\subsubsection{Proof of the Distributional View of Asymmetric Value Estimation}
\label{appen-sec:divl-proof}
We provide the proof of Proposition~\ref{prop:distributional_view} stated in Section~\ref{subsec:DIVL}. The goal is to show that, under idealized conditions, direct asymmetric optimization over dataset action-values and the two-step procedure of first fitting the state-conditioned distribution of dataset $Q$-values and then extracting the corresponding asymmetric statistic yield the same optimal scalar value.

Define the generalized asymmetric $L_p$ loss
\begin{equation}
\rho_{\tau, p}(u) = |\tau - \mathbb{I}(u < 0)| \cdot |u|^p,
\end{equation}
where $\tau\in(0,1)$ is the asymmetry parameter. In standard IQL, the scalar value is obtained by directly minimizing
\begin{equation}
J_{\text{direct}}(v) = \mathbb{E}_{\mathbf{a} \sim \mathcal{D}(\cdot\mid s)} \left[ \rho_{\tau, p}(Q(s, \mathbf{a}) - v) \right].
\end{equation}
The first-order optimality condition is
\begin{equation}
\frac{\mathrm{d}}{\mathrm{d}v} J_{\text{direct}}(v) = \int \mathcal{D}(\mathbf{a}\mid s) \cdot \frac{\mathrm{d}}{\mathrm{d}v} \rho_{\tau, p}(Q(s, \mathbf{a}) - v) \, \mathrm{d}\mathbf{a} = 0.
\end{equation}

Now consider DIVL in the idealized limit of infinitely fine discretization and sufficient model capacity. Let $p_\psi(z\mid s)$ denote the learned state-conditioned density over dataset $Q$-values. At optimum, the cross-entropy objective recovers the pushforward distribution induced by $\mathbf{a}\sim\mathcal{D}(\cdot\mid s)$ through the mapping $v = Q(s,\mathbf{a})$:
\begin{equation}
p_\psi(v\mid s) = P(v = Q(s, \mathbf{a}) \mid \mathbf{a}\sim \mathcal{D}(\cdot\mid s)).
\end{equation}
Thus, for any integrable test function $f(z)$,
\begin{equation}
\mathbb{E}_{v\sim p_\psi(\cdot\mid s)}[f(z)] = \mathbb{E}_{\mathbf{a}\sim\mathcal{D}(\cdot\mid s)}[f(Q(s,\mathbf{a}))].
\end{equation}
The second step of DIVL extracts a scalar statistic by minimizing
\begin{equation}
J_{\text{dist}}(v) = \mathbb{E}_{u \sim p_\psi(\cdot\mid s)} \left[ \rho_{\tau, p}(u - v) \right].
\end{equation}
Its first-order optimality condition is
\begin{equation}
\frac{\mathrm{d}}{\mathrm{d}v} J_{\text{dist}}(v) = \int p_\psi(u\mid s) \cdot \frac{\mathrm{d}}{\mathrm{d}v} \rho_{\tau, p}(u - v) \, \mathrm{d}u = 0.
\end{equation}
Because $p_\psi(\cdot\mid s)$ is exactly the pushforward of $\mathbf{a}\sim\mathcal{D}(\cdot\mid s)$ under the random variable $u = Q(s,\mathbf{a})$, the above integral is identical to the direct objective's optimality condition after change of variables. Therefore, $J_{\text{direct}}$ and $J_{\text{dist}}$ admit the same minimizer $v^*$ under the stated idealized assumptions.

This establishes that direct asymmetric optimization and the distribution-fit-then-extract procedure are equivalent in the limit. In particular, $p=2$ recovers the expectile statistic used in standard IQL, while $p=1$ recovers the quantile statistic used by DIVL.

\subsubsection{Analysis of Direct Backpropagation for Flow-Based Policy}
\label{appen-sec:flow-backprop}
Consider a flow-based policy that generates an action $x=x_1$ by integrating the vector field $\mathrm{d}{x}_t = f_\theta(x_t, t)$ from $t=0$ to $1$ starting from $x_0\sim\mathcal{N}$. Writing $x_1=x_1(x_0;\theta)$ for the terminal sample induced by the flow, the standard RL objective for reward fine-tuning is
\begin{equation}
J(\theta)=\mathbb{E}_{x_0\sim \mathcal{N}}\left[R\big(x_1(x_0;\theta)\big)\right],
\end{equation}
and a vanilla policy gradient requires differentiating through the entire ODE trajectory:
\begin{equation}
\nabla_\theta J(\theta) = \mathbb{E}_{x_0 \sim \mathcal{N}} \left[ \nabla_x R(x_1) \cdot \int_{0}^{1} \Phi(1, t) \frac{\partial f_\theta(x_t, t)}{\partial \theta} dt \right],
\end{equation}
where $\Phi(1, t) = \frac{\partial x_1}{\partial x_t}$ is the sensitivity matrix along the flow. In practice, this formulation is computationally expensive and numerically fragile because it requires backpropagation through the full ODE solver~\citep{domingo2024adjoint}. Adjoint Matching (Section~\ref{subsec:AM}) avoids this issue by reformulating trajectory-level optimization as local regression targets along the flow path.

\subsection{Implementation and Training Details}

\subsubsection{Offline Data}
\label{appen-sec:data}
The offline buffer $\mathcal{B}_{\mathrm{off}}$ consists of three types of data: \emph{demonstration} data collected by human experts, \emph{rollout} data produced by historical policies during prior evaluations, and \emph{play} data in which a human operator explores failure modes and edge cases. Demonstrations are successful trajectories, rollouts contain both successes and failures, and play data is treated as unsuccessful exploratory data. Table~\ref{tab:data-composition} summarizes the data composition in hours by task. Fig.~\ref{fig:data-composition-pie} shows the aggregate source distribution, illustrating the relative contribution of demonstrations, historical rollouts, and play data.

\begin{table*}[t]
\centering
\caption{\textbf{Offline data composition (hours).} Demonstrations are expert-collected successful data; rollouts are generated by historical policies and contain both successes and failures; play data consists of human-guided explorations of failure modes.}
\label{tab:data-composition}
\setlength{\tabcolsep}{5pt}
\small
\begin{tabular}{l cccc c}
\toprule
\textbf{Task} & \textbf{Demo} & \textbf{Rollout (Succ)} & \textbf{Rollout (Fail)} & \textbf{Play} & \textbf{Total} \\
\midrule
Restocking       & 14.5 & 10.7 & 1.7 & 7.5 & 34.4 \\
Correction       & 12.7 & 10.8 & 1.7 & 9.7 & 34.9 \\
Freezer          & 10.8 & 7.7  & 4.6 & 3.3 & 26.4 \\
Open-Cooler      & 11.1 & 13.7 & 1.3 & 0.8 & 26.9 \\
\rowcolor[gray]{0.93} \textit{Grocery Restocking Tasks (subtotal)} & \textit{49.2} & \textit{42.9} & \textit{9.3} & \textit{21.3} & \textit{122.7} \\
\midrule
Gongfu Tea      & 102.3 & 12.4 & 4.4  & 43.6 & 162.7 \\
Fruit Juicing    & 100.5 & 17.4 & 14.7 & 47.1 & 179.8 \\
Cocktail         & 47.3  & 4.4  & 7.4  & 28.0 & 87.1  \\
Shoebox          & 37.3  & 11.7 & 3.3  & 48.0 & 100.3 \\
\rowcolor[gray]{0.93} \textit{Long-Horizon Tasks (subtotal)} & \textit{287.5} & \textit{45.9} & \textit{29.8} & \textit{166.7} & \textit{529.8} \\
\midrule
\rowcolor[gray]{0.85} \textbf{All Tasks} & \textbf{336.6} & \textbf{88.8} & \textbf{39.2} & \textbf{187.9} & \textbf{652.5} \\
\bottomrule
\end{tabular}
\end{table*}


\providecolor{finchgreen1}{HTML}{D0DEDC}
\providecolor{finchgreen2}{HTML}{B7CCC8}
\providecolor{finchgreen3}{HTML}{9DBAB5}
\providecolor{finchgreen5}{HTML}{688E87}
\providecolor{finchgreen6}{HTML}{577872}
\providecolor{finchred1}{HTML}{FDDED9}
\providecolor{finchred2}{HTML}{EAC4BF}
\providecolor{finchred3}{HTML}{D5A09D}
\providecolor{finchred4}{HTML}{BD7F76}
\providecolor{finchred5}{HTML}{B46655}
\providecolor{finchred6}{HTML}{80422F}

\begin{figure*}[t]
\centering
\begin{minipage}[c]{0.48\textwidth}
\centering
\begin{tikzpicture}
  \fill[finchgreen2] (0,0) -- ( 90.00:2.2) arc ( 90.00:108.98:2.2) -- cycle; 
  \fill[finchgreen3] (0,0) -- (108.98:2.2) arc (108.98:128.23:2.2) -- cycle; 
  \fill[finchgreen5] (0,0) -- (128.23:2.2) arc (128.23:142.80:2.2) -- cycle; 
  \fill[finchgreen6] (0,0) -- (142.80:2.2) arc (142.80:157.64:2.2) -- cycle; 
  \fill[finchred2]   (0,0) -- (157.64:2.2) arc (157.64:247.41:2.2) -- cycle; 
  \fill[finchred3]   (0,0) -- (247.41:2.2) arc (247.41:346.61:2.2) -- cycle; 
  \fill[finchred5]   (0,0) -- (346.61:2.2) arc (346.61:394.66:2.2) -- cycle; 
  \fill[finchred6]   (0,0) -- (394.66:2.2) arc (394.66:450.00:2.2) -- cycle; 
  \foreach \a in {90.00, 108.98, 128.23, 142.80, 157.64, 247.41, 346.61, 394.66} {
    \draw[white, line width=0.8pt] (0,0) -- (\a:2.2);
  }
  \node[white, font=\footnotesize\bfseries] at (202.52:1.36) {24.9\%};
  \node[white, font=\footnotesize\bfseries] at (297.01:1.36) {27.6\%};
  \node[white, font=\footnotesize\bfseries] at (370.63:1.36) {13.3\%};
  \node[white, font=\footnotesize\bfseries] at (422.33:1.36) {15.4\%};
\end{tikzpicture}

\vspace{0.4em}
{\footnotesize
\setlength{\tabcolsep}{3pt}
\begin{tabular}{@{}cl@{\hspace{1.2em}}cl@{}}
\tikz\fill[finchgreen2] (0,0) rectangle (0.7em,0.7em); & Restocking\,(5.3\%) &
\tikz\fill[finchred2]   (0,0) rectangle (0.7em,0.7em); & Gongfu Tea\,(24.9\%) \\
\tikz\fill[finchgreen3] (0,0) rectangle (0.7em,0.7em); & Correction\,(5.3\%) &
\tikz\fill[finchred3]   (0,0) rectangle (0.7em,0.7em); & Fruit Juicing\,(27.6\%) \\
\tikz\fill[finchgreen5] (0,0) rectangle (0.7em,0.7em); & Freezer\,(4.0\%) &
\tikz\fill[finchred5]   (0,0) rectangle (0.7em,0.7em); & Cocktail\,(13.3\%) \\
\tikz\fill[finchgreen6] (0,0) rectangle (0.7em,0.7em); & Open-Cooler\,(4.1\%) &
\tikz\fill[finchred6]   (0,0) rectangle (0.7em,0.7em); & Shoebox\,(15.4\%) \\
\end{tabular}
}

\vspace{0.3em}
{\footnotesize (a) By task: \textcolor{finchgreen6}{the Grocery Restocking tasks} 18.8\% \textbar{} \textcolor{finchred6}{Long-Horizon} 81.2\%}
\end{minipage}
\hfill
\begin{minipage}[c]{0.48\textwidth}
\centering
\begin{tikzpicture}
  \fill[finchgreen5] (0,0) -- ( 90.00:2.2) arc ( 90.00:275.71:2.2) -- cycle; 
  \fill[finchgreen3] (0,0) -- (275.71:2.2) arc (275.71:324.70:2.2) -- cycle; 
  \fill[finchred3]   (0,0) -- (324.70:2.2) arc (324.70:346.33:2.2) -- cycle; 
  \fill[finchred5]   (0,0) -- (346.33:2.2) arc (346.33:450.00:2.2) -- cycle; 
  \foreach \a in {90.00, 275.71, 324.70, 346.33} {
    \draw[white, line width=0.8pt] (0,0) -- (\a:2.2);
  }
  \node[white, font=\footnotesize\bfseries] at (182.86:1.36) {51.6\%};
  \node[white, font=\footnotesize\bfseries] at (300.21:1.36) {13.6\%};
  \node[white, font=\footnotesize\bfseries] at (398.17:1.36) {28.8\%};
  \draw[gray!70, line width=0.4pt] (335.52:2.2) -- (335.52:2.55);
  \node[font=\footnotesize, anchor=west] at (335.52:2.6) {6.0\%};
\end{tikzpicture}

\vspace{0.4em}
{\footnotesize
\setlength{\tabcolsep}{3pt}
\begin{tabular}{@{}cl@{}}
\tikz\fill[finchgreen5] (0,0) rectangle (0.7em,0.7em); & Demonstration\,(51.6\%, 336.6\,h) \\
\tikz\fill[finchgreen3] (0,0) rectangle (0.7em,0.7em); & Rollout, success\,(13.6\%, \phantom{0}88.8\,h) \\
\tikz\fill[finchred3]   (0,0) rectangle (0.7em,0.7em); & Rollout, failure\,(6.0\%, \phantom{0}39.2\,h) \\
\tikz\fill[finchred5]   (0,0) rectangle (0.7em,0.7em); & Play\,(28.8\%, 187.9\,h) \\
\end{tabular}
}

\vspace{0.3em}
{\footnotesize (b) By source, colored by outcome: \textcolor{finchgreen6}{Successful} 65.2\% \textbar{} \textcolor{finchred6}{Failure} 34.8\%}
\end{minipage}

\caption{\textbf{Offline data composition of the 652.5-hour buffer along two axes.} \textbf{(a)} Distribution across tasks: the grocery restocking tasks (\textcolor{finchgreen6}{green}) and long-horizon tasks (\textcolor{finchred6}{red}); long-horizon episodes dominate the buffer by volume due to their substantially longer duration. \textbf{(b)} Distribution across the three data sources---expert \emph{demonstrations} (always successful), \emph{rollouts} from historical policies (mixed successful and failure outcomes), and human-guided failure-mode \emph{play} (always unsuccessful). Wedges are colored by trajectory outcome so that the overall success/failure split across the buffer is directly legible: roughly one-third of the buffer is failure data, which the behavior-cloning baselines cannot use but which provides an informative learning signal for LWD. Per-task hours are reported in Table~\ref{tab:data-composition}.}
\label{fig:data-composition-pie}
\end{figure*}
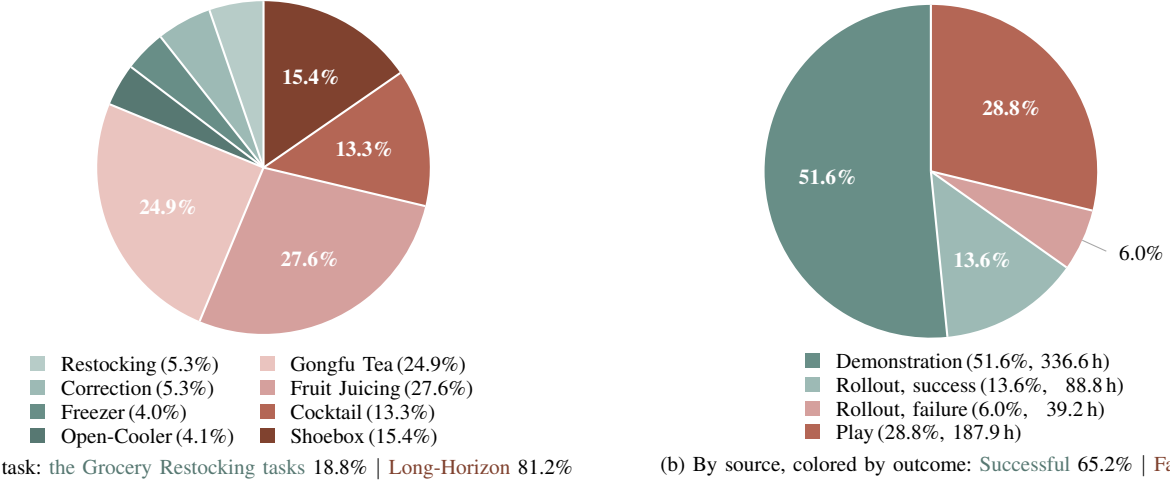

\subsubsection{Training Hyperparameters}
\label{appen-sec:training-details}
The policy emits action chunks with horizon $H=30$. The policy is optimized with AdamW~\citep{loshchilov2018decoupled} using a base learning rate of \(2\times10^{-5}\) and a cosine decay schedule. The value and critic networks are trained with Adam using a base learning rate of \(5\times10^{-4}\), also with a cosine decay schedule.

\begin{figure}[htbp]
    \centering
    \includegraphics[width=0.98\linewidth]{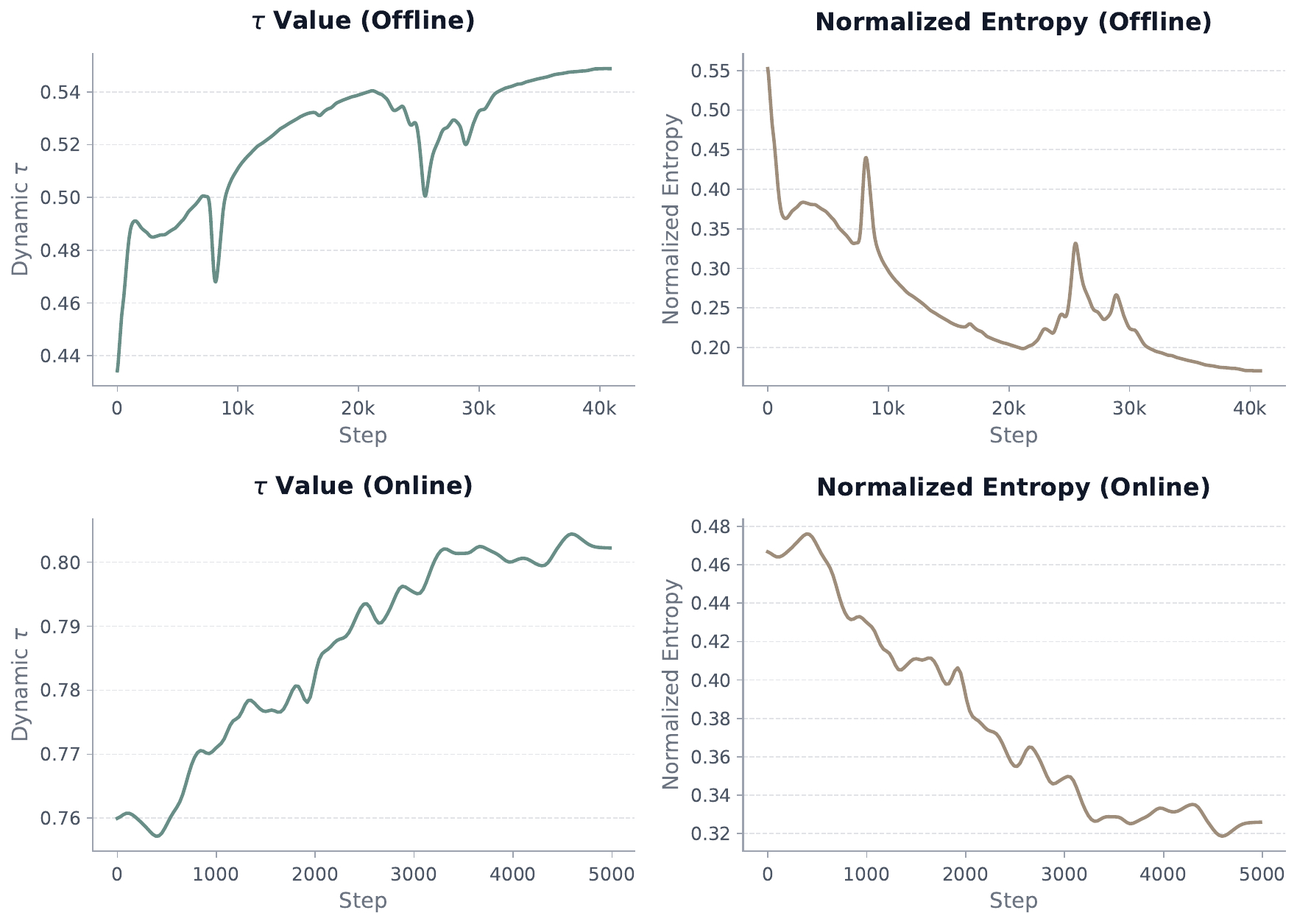}
    \caption{\textbf{Dynamic $\tau$ and normalized entropy during offline-to-online training.} All curves are smoothed for readability. Entropy decreases throughout both stages, indicating increasing confidence in value estimation. Accordingly, $\tau$ is increased, leading to improved training performance.
    }
    \label{fig:app_dynamic_tau_vis}
\end{figure}

For temporal-difference backups, we use \(\gamma=0.9999\). During offline training, we use \(\tau_{\text{base}}=0.6\) and uncertainty-sensitivity coefficient \(\alpha=0.3\) for DIVL. During online training, we use \(\tau_{\text{base}}=0.9\) and \(\alpha=0.3\). Target critic and value networks are updated with EMA rate \(0.005\), and the QAM policy-extraction temperature is \(\lambda=2\).
The $\tau$ and entropy values during offline and online training are visualized in Fig.~\ref{fig:app_dynamic_tau_vis}.
Entropy decreases throughout offline-to-online training, indicating increasing confidence of value functions.
Accordingly, the expectile parameter $\tau$ is increased, encouraging the policy to favor higher-value solutions.

\begin{figure}[htbp]
    \centering
    \includegraphics[width=0.98\linewidth]{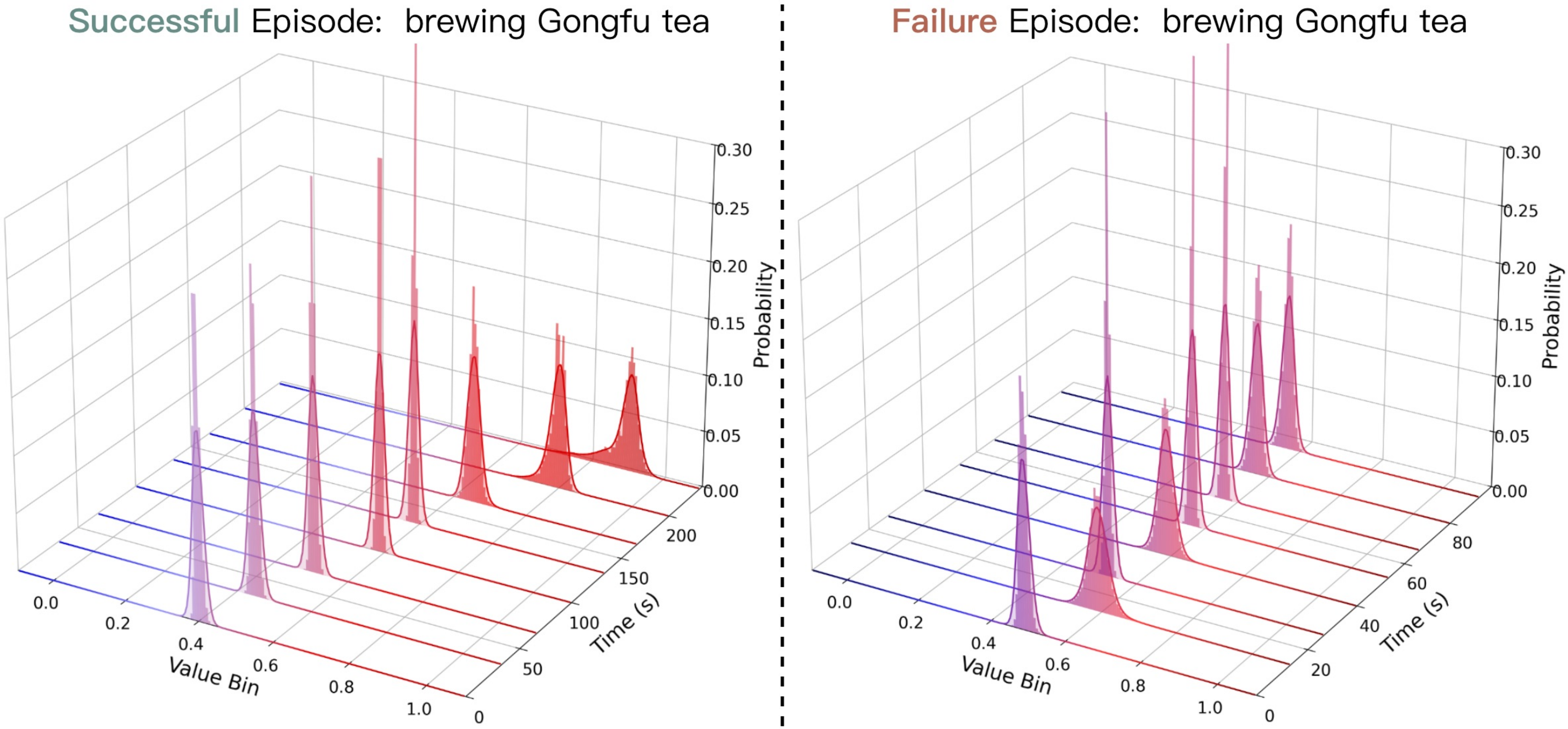}
    \caption{\textbf{Predicted Value Distributions}. In the successful episode, the predicted distribution remains unimodal and its mode increases steadily from approximately 0.4 to 1.0. In contrast, the failure episode shows limited mode progression, rising only from approximately 0.5 to 0.6 before plateauing.
    }
    \label{fig:value_dist_vis}
\end{figure} 

For value learning, offline training uses 10-step chunk-level TD for long-horizon tasks and 1-step chunk-level TD for the grocery restocking tasks. Online training uses 1-step chunk-level TD for all tasks. During online training, each learner update samples mini-batches from \(\mathcal{B}_{\text{off}} \cup \mathcal{B}_{\text{on}}\) with an approximately balanced ratio of \(1{:}1\).

\subsubsection{Checkpoint Initialization}
We first train an imitation-learning checkpoint by adapting the pretrained $\pi_{0.5}$ VLA policy on the demonstration data with behavior cloning. LWD (Offline) initializes its policy from this imitation-learning checkpoint, then trains the policy with the Adjoint Matching loss and trains the critic and distributional value model with DIVL. LWD (Online) initializes from the LWD (Offline) checkpoint, including both policy and value-learning modules, and continues training on mixed offline-online replay.

\subsection{Additional Experimental Details}

\subsubsection{Reference Policy and Baseline Implementations}
\label{appen-sec:experiment-details}
We obtain the reference policy by supervised fine-tuning ~\citep{lipman2022flow} the pretrained $\pi_{0.5}$ VLA policy on 336.6 hours of demonstration data, as shown in Table~\ref{tab:data-composition}. The model is trained with a flow-matching loss, where the interpolated noisy action \( \mathbf{a}^w \) is defined in Eq.~\eqref{eq:fm-interpolation}.The objective is to train conditional vector field $f_\theta(s,\mathbf{a}^w,w)$ to match the velocity $\mathbf{a}^1-\mathbf{a}^0$ and minimizes:
\begin{equation}
    \mathcal{L}_{\mathrm{SFT}} = \mathbb{E} \left[ \left\| f_\theta(s, \mathbf{a}^w, w) - (\mathbf{a}^1 - \mathbf{a}^0) \right\|_2^2 \right],
\end{equation} And this reference policy is used for all the post-training methods.

For the RECAP~\citep{pi06} baseline, we initialize from the reference policy and adapt RECAP to the eight-task generalist setting. We collect two rounds of autonomous rollouts: Round 1 uses the SFT checkpoint, and Round 2 uses the RECAP checkpoint obtained after training on Round 1. Each round contains approximately 60 robot-hours pooled across all eight tasks. Following RECAP, we train a value model to compute advantage labels over the combined dataset of demonstrations and both autonomous rollout rounds; the value model uses the same value-network architecture as LWD, described in Section~\ref{subsec:network-architecture}, but only the value head. We compute the lookahead advantage and binary improvement label as
\begin{equation}
A(s_t,\mathbf{a}_t)
=
\sum_{t'=t}^{t+H-1} r_{t'}
+
V(s_{t+H})
-
V(s_t),
\end{equation}
\begin{equation}
I_t =
\mathbbm{1}\!\left[
A^{\pi_{\mathrm{ref}}}(s_t,\mathbf{a}_t) > \epsilon
\;\lor\;
c_t = 1
\right],
\end{equation}
where \(c_t\) indicates that the transition is a human intervention or correction, which is treated as positive following RECAP when present. We use \(H=30\) as our action horizon length and select a single global advantage threshold \(\epsilon\) so that 30\% of transitions in the combined training set satisfy the positive-advantage condition. This threshold is selected from training data only and is shared across all tasks to avoid task-specific tuning. After the second rollout round, we train RECAP for one epoch over the combined dataset and evaluate the resulting checkpoint.

For the HG-DAgger~\citep{kelly2019hg} baseline, we initialize from the same reference policy checkpoint and run interactive imitation learning on the eight-task suite. During online execution, human operators provided intervention segments when corrections are needed. These intervention segments are aggregated with autonomous rollouts to form an online training buffer of approximately 60 robot-hours pooled across all eight real-world tasks. The online buffer, together with the offline demonstration data buffer, is used for the training of the HG-DAgger method. We train HG-DAgger from the reference policy checkpoint using the same batch size and training-time budget as the corresponding online post-training runs, and evaluate the resulting checkpoint.

For a fair comparison, the post-training baselines use the same policy optimizer and learning-rate schedule as LWD.

\subsubsection{Complete Value-Estimation Ablation Results}
Table~\ref{tab:ablation_value_estimation} reports the complete per-task results for the value-estimation ablation summarized in Section~\ref{sec:exp}. The comparison isolates the value-learning method by replacing DIVL with scalar expectile value regression while keeping the remaining training setup fixed.

\begin{table*}[t]
\centering
\caption{\textbf{Ablation of value learning design (complete results).}
Complete results on grocery restocking tasks and long-horizon tasks. We compare continuous expectile regression and our distributional implicit value learning under offline and online settings. We report task success rate for each task and the average across all eight tasks. The best result per column is shown in bold.}
\label{tab:ablation_value_estimation}
\setlength{\tabcolsep}{3.5pt}
\renewcommand{\arraystretch}{1.12}
\small
\resizebox{\textwidth}{!}{%
\begin{tabular}{llccccccccc}
\toprule
\multicolumn{2}{c}{\multirow{2}{*}{\textbf{Method}}} &
\multicolumn{4}{c}{\textbf{Grocery Restocking Tasks}} &
\multicolumn{4}{c}{\textbf{Long-Horizon Tasks}} &
\multirow{2}{*}{\textbf{Average}} \\
\cmidrule(lr){3-6}\cmidrule(lr){7-10}
\multicolumn{2}{c}{} & 
\textbf{Restocking} &
\textbf{Correction} &
\textbf{Freezer} &
\textbf{Open-Cooler} &
\textbf{Gongfu Tea} &
\textbf{Fruit Juice} &
\textbf{Cocktail} &
\textbf{Shoebox} &
\\
\midrule
\multirow{2}{*}{Offline}
& Expectile Regression
& 0.85 & \textbf{1.00} & \textbf{1.00} & \textbf{1.00}
& 0.68 & 0.74 & 0.71 & 0.76 & 0.84 \\
& DIVL (ours)
& \textbf{1.00} & \textbf{1.00} & 0.92 & 0.95
& 0.72 & 0.74 & 0.83 & 0.86 & 0.88 \\
\midrule
\multirow{2}{*}{Online}
& Expectile Regression
& 0.95 & \textbf{1.00} & 0.92 & \textbf{1.00}
& 0.76 & 0.76 & 0.77 & 0.84 & 0.88 \\
& \textbf{DIVL (ours)}
& \textbf{1.00} & \textbf{1.00} & 0.97 & 0.98
& \textbf{0.89} & \textbf{0.90} & \textbf{0.93} & \textbf{0.92} & \textbf{0.95} \\
\bottomrule
\end{tabular}%
}
\end{table*}

\subsubsection{Complementary Qualitative Results of DIVL} \label{appen-sec:value_dist_vis}
Fig.~\ref{fig:value_dist_vis} visualizes the predicted value distributions for the same episodes shown in Fig.~\ref{fig:value_vis}. In the successful episode, the predicted distribution remains unimodal, with its mode steadily increasing from approximately 0.4 to 1.0 as the task progresses. In contrast, the failure episode exhibits only marginal mode progression, increasing from approximately 0.5 to 0.6 before plateauing. These results indicate that the predicted value distribution provides a fine-grained signal to track policy progress and distinguish successful execution from failure cases.


\subsection{Distributed Data Infrastructure}
\label{appen-sec:lwd-infra}

\begin{figure}[t]
    \centering
    \resizebox{\linewidth}{!}{%
    \begin{tikzpicture}[
        node distance=0.8cm and 1.2cm,
        box/.style={rectangle, draw, rounded corners, minimum height=0.9cm, minimum width=1.8cm, align=center, font=\small},
        actor/.style={box, fill=blue!15},
        distservice/.style={box, fill=orange!20},
        coord/.style={box, fill=orange!40, very thick},
        learner/.style={box, fill=green!15},
        readerbox/.style={box, fill=green!30, very thick},
        arrow/.style={-{Stealth[length=2mm]}, thick},
        dashedarrow/.style={-{Stealth[length=2mm]}, thick, dashed},
        label/.style={font=\footnotesize\itshape, text=gray}
    ]

    \node[actor] (robot1) {Robot 1};
    \node[actor, below=0.4cm of robot1] (robot2) {Robot 2};
    \node[below=0.3cm of robot2, font=\small] (dots1) {$\vdots$};
    \node[actor, below=0.3cm of dots1] (robotN) {Robot $N$};
    \node[actor, right=0.8cm of robot2] (edgeclient) {Edge\\Client};

    \node[distservice, right=1.5cm of edgeclient, yshift=1.2cm] (oss) {Object\\Storage};
    \node[distservice, right=1.5cm of edgeclient, yshift=-1.2cm] (mq) {Message\\Queue};
    \node[coord, right=2.0cm of oss, yshift=-1.2cm] (coordinator) {Coordinator};

    \node[readerbox, right=3.0cm of coordinator, yshift=0.7cm] (reader1) {DRB\\Reader};
    \node[learner, right=0.9cm of reader1] (cl1) {Cloud\\Learner};
    \node[readerbox, below=0.7cm of reader1] (readerM) {DRB\\Reader};
    \node[learner, right=0.9cm of readerM] (clM) {Cloud\\Learner};

    \begin{scope}[on background layer]
        \node[draw=blue!50, dashed, rounded corners, fit=(robot1)(robotN)(edgeclient), inner sep=0.3cm] (actorbox) {};
        \node[draw=orange!50, dashed, rounded corners, fit=(oss)(mq)(coordinator), inner sep=0.3cm] (distservicebox) {};
        \node[draw=green!50, dashed, rounded corners, fit=(reader1)(cl1)(readerM)(clM), inner sep=0.3cm] (learnerbox) {};
    \end{scope}

    \node[font=\small\bfseries, above=0.1cm of actorbox] {Actor Fleet};
    \node[font=\small\bfseries, above=0.1cm of distservicebox] {Distribution \& Coordination};
    \node[font=\small\bfseries, above=0.1cm of learnerbox] {Cloud Learner (multi-host SPMD JAX)};

    \draw[arrow] (robot1.east) -- (edgeclient.west);
    \draw[arrow] (robot2.east) -- (edgeclient.west);
    \draw[arrow] (robotN.east) -- (edgeclient.west);

    \draw[arrow] (edgeclient.east) -- ++(0.4,0) |- node[label, pos=0.25, above]{episodes} (oss.west);
    \draw[arrow] (edgeclient.east) -- ++(0.4,0) |- node[label, pos=0.25, below]{events} (mq.west);

    \draw[arrow] (mq.east) -| node[label, pos=0.25, above]{notify} (coordinator.south);

    \draw[arrow] (coordinator.east) -- node[label, pos=0.4, above]{versioned snapshot} (reader1.west);
    \draw[arrow] (coordinator.east) -- (readerM.west);


    \draw[arrow] (reader1.east) -- node[label, above]{batch} (cl1.west);
    \draw[arrow] (readerM.east) -- (clM.west);

    \coordinate (mqdown) at ($(mq.south) + (0,-2.5)$);
    \coordinate (fanout) at ($(robot2.west) + (-0.6,0)$);
    \coordinate (fanout_left) at ($(fanout) + (-1,0)$);
    \coordinate (leftbottom) at (fanout_left |- mqdown);

    \draw[dashedarrow, blue!60] (clM.south) -- ++(0,-0.7) -| node[label, pos=0.25, below, text=blue!60]{model weights} (mq.south);

    \draw[dashedarrow, blue!60] (mq.south) -- (mqdown) -- (leftbottom) -- (fanout_left) -- (fanout);

    \draw[dashedarrow, blue!60] (fanout) -- (robot2.west);
    \draw[dashedarrow, blue!60] (fanout) |- (robot1.west);
    \draw[dashedarrow, blue!60] (fanout) |- (robotN.west);

    \node[font=\scriptsize\itshape, text=blue!60, above left=0.1cm of fanout] {fanout};
    \end{tikzpicture}%
    }
    \caption{\textbf{Distributed data infrastructure for LWD.} Robot actors upload episodes to object storage and publish event notifications to a message queue. A central \emph{Coordinator} consumes notifications, fetches episode metadata, and commits versioned snapshots. The learner runs as a multi-host SPMD JAX program; on each node, the dataset (\emph{DRB Reader}) holds a snapshot-bound view, spawns a prefetcher subprocess to download payloads from object storage, and feeds mini-batches to the local learner process. All DRB Readers synchronize on the same snapshot via a cross-host barrier. Updated model parameters produced by the collective are published back to all robot actors via the message-queue-backed publish-subscribe channel.}
    \label{fig:lwd-infra}
\end{figure}
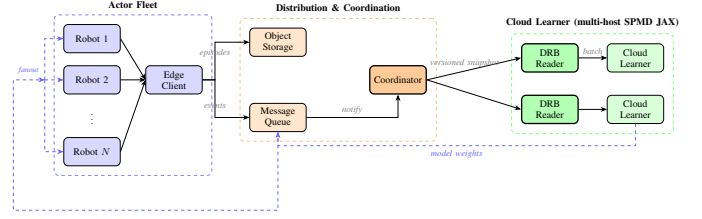

Fig.~\ref{fig:lwd-infra} illustrates LWD's training data infrastructure, which links a fleet of robot actors to a multi-host learner via a versioned-snapshot data plane. On the actor side, each robot runs an edge client that accumulates per-frame observations into complete episodes and uploads them to distributed object storage at episode boundaries; episode metadata is persisted by a business service and event notifications are published to a message queue.

On the cloud side, a central \emph{Coordinator} consumes event notifications from the message queue, fetches episode metadata from object storage, and commits monotonically increasing snapshot versions that define the training data view at each step. The learner runs as a multi-host SPMD JAX program, with one process per node driving all local accelerators. Each process instantiates a \emph{Distributed Replay Buffer (DRB) Reader} as its dataset; before each training step, all DRB Readers synchronize on the same snapshot version via a cross-host barrier, ensuring the SPMD collective sees a globally consistent dataset view despite asynchronous edge ingestion. Each DRB Reader spawns a prefetcher subprocess that downloads payloads from object storage in parallel; placing one prefetcher per node is sufficient to saturate the per-node read bandwidth available from the underlying distributed filesystem in our deployment.

Model parameters produced by the SPMD collective are published to a publish-subscribe channel that fans out to all robot actors, which reload the new policy at episode boundaries. Across the entire design, the Coordinator is the only orchestration singleton; both the actor fleet and the learner scale independently.

We characterize this infrastructure along two operational axes that are critical for online RL: whether every collected episode is reliably incorporated into training, and how quickly new data and updated policies traverse the actor--learner loop.

\subsubsection{End-to-End Reliability}

The system provides at-least-once end-to-end delivery for every episode produced on the actor side. (i)~Object-storage uploads commit atomically (readers see either the fully-uploaded payload or no object) and are retried until persisted. (ii)~Episode metadata is committed via a transactional insert in the business service, then announced to a durable message queue with delivery acknowledgment, so notifications survive coordinator restarts. (iii)~Per-node prefetcher download tasks are requeued on failure with bounded retries; on snapshot commit, the snapshot data and the version pointer are updated atomically, so partial failures cannot leave a snapshot inconsistent. In our profiled 8-hour, 16-actor run of 1{,}604 episodes, every episode ingested in steady state completed the full end-to-end pipeline.

\begin{table}[t]
\centering
\caption{\textbf{Operational Latency.} End-to-end latency measured on the same 8-hour, 16-actor online-RL run as the End-to-End Reliability subsection above. Absolute values are sensitive to network configuration and link contention and may vary across deployments.}
\label{tab:lwd-metrics}
\small
\begin{tabular}{lrr}
\toprule
\textbf{Path} & \textbf{P50} & \textbf{P99} \\
\midrule
Episode produced $\to$ available to learner   & 41\,s   & 148\,s \\
Model published $\to$ received by actor       & 38\,s   & 55\,s  \\
\bottomrule
\end{tabular}
\end{table}

\subsubsection{Operational Latency}

We report the two end-to-end latencies that govern the tightness of the actor-learner loop: (i)~\emph{episode-to-learner}: the elapsed time from when an episode is produced on an actor to when it becomes available for the learner to sample; and (ii)~\emph{model-to-actor}: the elapsed time from when the learner publishes a new policy to when the actor has loaded it for the next rollout. Table~\ref{tab:lwd-metrics} reports both on the same 8-hour, 16-actor run as the End-to-End Reliability subsection above. Both latencies are dominated by object-storage I/O on the actor-to-cloud link --- the episode payload in one direction, the policy artifact in the other --- so absolute values are sensitive to link bandwidth and contention and may vary substantially across deployments.

\end{document}